\documentclass{article}

\usepackage{fullpage}

\newif\ifcomment\commentfalse
\usepackage{algorithmic}
\usepackage{algorithm}
\usepackage{amsfonts}
\usepackage{amsmath}
\usepackage{amssymb}
\usepackage{bbm}
\usepackage{bm}
\usepackage{color}
\usepackage{comment}
\usepackage{epsfig}
\usepackage{exscale}
\usepackage{framed}
\usepackage{graphicx}
\usepackage{rotating}
\usepackage{latexsym}
\usepackage{mdwlist}
\usepackage{microtype}
\usepackage{multirow}
\usepackage{subfigure}
\usepackage[font=small,labelfont=bf,tableposition=top]{caption}
\usepackage{times}
\usepackage{url}
\usepackage[abs]{overpic}
\usepackage{pict2e}

\newcommand{\explain}[2]{\underbrace{#2}_{\mbox{\footnotesize{#1}}}}
\newcommand{\dir}[1]{\mbox{Dir}(#1)}
\newcommand{\mult}[1]{\mbox{Mult}( #1)}
\newcommand{\g}{\, | \,}

\newcommand{\LG}[1]{\log \Gamma \left( \textstyle #1 \right)}

\newcommand{\digam}[1]{\Psi \left( \textstyle #1 \right) }
\newcommand{\ddigam}[1]{\Psi' \left( \textstyle #1 \right) }
\newcommand{\e}[2]{\mathbb{E}_{#1}\left[ #2 \right] }

\newcommand{\elbo}{\mathcal{L}}

\newcommand{\hidetext}[1]{ }
\ifcomment
\newcommand{\jbgcomment}[1]{  \colorbox{red}{   \parbox{.8\linewidth}{ JBG: #1}  }}
\newcommand{\yhcomment}[1]{  \colorbox{green}{  \parbox{.8\linewidth}{ YH:  #1}  }}
\newcommand{\reviewercomment}[1]{  \colorbox{blue}{  \parbox{.8\linewidth}{Reviewer:  #1}  }}
\newcommand{\brscomment}[1]{  \colorbox{blue}{  \parbox{.8\linewidth}{BRS:  #1}  }}
\newcommand{\zkcomment}[1]{  \colorbox{cyan}{  \parbox{.8\linewidth}{ZK:  #1}  }}
\newcommand{\ctcomment}[1]{
  \colorbox{blue}{  \parbox{.8\linewidth}{CT:  #1}  }}
\newcommand{\swcomment}[1]{ \colorbox{yellow}{ \parbox{.8\linewidth}{ SW: #1}  }}
\else
\newcommand{\jbgcomment}[1]{ }
\newcommand{\yhcomment}[1]{ }
\newcommand{\brscomment}[1]{ }
\newcommand{\reviewercomment}[1]{ }
\newcommand{\zkcomment}[1]{ }
\newcommand{\ctcomment}[1]{ }
\newcommand{\swcomment}[1]{ }
\fi

\definecolor{grey}{rgb}{0.95,0.95,0.95}

\begin{document}
\title{Using Variational Inference and MapReduce to Scale Topic Modeling}


\author{{\bf Ke Zhai} \\
Computer Science \\
University of Maryland \\
College Park, MD, USA \\
zhaike@cs.umd.edu
\and
{\bf Jordan Boyd-Graber} \\
iSchool and UMIACS \\
University of Maryland \\
College Park, MD, USA \\
jbg@umiacs.umd.edu
\and
{\bf Nima Asadi} \\
Computer Science \\
University of Maryland \\
College Park, MD, USA \\
nima@cs.umd.edu
}


%Telephone: (800) 555--1212, Fax: (888) 555--1212}


\maketitle
\begin{abstract}
  Latent Dirichlet Allocation (LDA) is a popular topic modeling technique for
  exploring document collections.  Because of the increasing prevalence of large
  datasets, there is a need to improve the scalability of inference of LDA.  In
  this paper, we propose a technique called ~\emph{MapReduce LDA} (Mr. LDA) to
  accommodate very large corpus collections in the MapReduce framework.  In
  contrast to other techniques to scale inference for LDA, which use Gibbs
  sampling, we use variational inference. Our solution efficiently distributes
  computation and is relatively simple to implement. More importantly, this
  variational implementation, unlike highly tuned and specialized
  implementations, is easily extensible.  We demonstrate two extensions of the
  model possible with this scalable framework: informed priors to guide topic
  discovery and modeling topics from a multilingual corpus.
\end{abstract}

%\begin{IEEEkeywords}
%topic modeling; latent dirichlet allocation; variational inference; bayesian; scalable; mapreduce
%\end{IEEEkeywords}

% \fi
%\IEEEpeerreviewmaketitle

\jbgcomment{Ke: please check to see if driver, reducer, mapper, etc. are
  consistently capitalized (or not, whichever you prefer)}

\section{Introduction}

Because data from the web are big and noisy, algorithms that process large
document collections cannot solely depend on human annotations. One popular
technique for navigating large unannotated document collections is topic
modeling, which discovers the themes that permeate a corpus. Topic modeling is
exemplified by~\emph{Latent Dirichlet Allocation} (LDA), a generative model for
document-centric corpora~\cite{blei-03}. It is appealing for noisy data because
it requires no annotation and discovers, without any supervision, the thematic
trends in a corpus. In addition to discovering which topics exist in a corpus,
LDA also associates documents with these topics, revealing previously unseen
links between documents and trends over time.  Although our focus is on text
data, LDA is also widely used in the computer vision~\cite{fergus-05,wang-09b},
biology~\cite{airoldi-08,populationstructure}, and computational
linguistics~\cite{boyd-graber-09, griffiths-05} communities.

In addition to being noisy, data from the web are big. The MapReduce framework
for large-scale data processing~\cite{dean-04} is simple to learn but flexible
enough to be broadly applicable. Designed at Google and open-sourced by Yahoo,
MapReduce is one of the mainstays of industrial data processing and has also
been gaining traction for problems of interest to the academic community such as
machine translation~\cite{dyer-08}, language modeling~\cite{brants-07}, and
grammar induction~\cite{cohen-09}.

In this paper, we propose a parallelized LDA algorithm in MapReduce programming
framework (\emph{Mr. LDA}).  Mr. LDA relies on variational inference, as opposed
to the prevailing trend of using Gibbs sampling, which we argue is an effective
means of scaling out LDA in Section~\ref{sec:scaling}. Section~\ref{sec:model}
describes how variational inference fits naturally into the MapReduce
framework. In Section~\ref{sec:flexibility}, we discuss two specific extensions
of LDA to demonstrate the flexibility of the proposed framework.  These are an
informed prior to guide topic discovery and a new inference technique for
discovering topics in multilingual corpora~\cite{mimno-09}.  Next, we evaluate
Mr. LDA's ability to scale both in the number of documents and the number of
topics in Section~\ref{sec:exp:scaling} before concluding with
Section~\ref{sec:conc}.

\section{Scaling out LDA}
\label{sec:scaling}

In practice, probabilistic models work by maximizing the log-likelihood of 
observed data given the structure of an assumed probabilistic model. Less
technically, generative models tell a story of how your data came to be with
some pieces of the story missing; inference fills in the missing pieces with the
best explanation of the missing variables. Because exact inference is often
intractable (as it is for LDA), complex models require approximate inference.

\subsection{Why not Gibbs Sampling?}
\label{sec:scaling:gibbs}

One of the most widely used approximation techniques for such models is Markov
chain Monte Carlo (MCMC) sampling, where one samples from a Markov chain whose
limiting distribution is the posterior of interest~\cite{neal-93,robert-04}.
Gibbs sampling, where the Markov chain is defined by the conditional
distribution of each latent variable, has found widespread use in Bayesian
models~\cite{neal-93,teh-06b,griffiths-04,finkel-07}.  MCMC is a powerful
methodology, but it has drawbacks. Convergence of the sampler to its stationary
distribution is difficult to diagnose, and sampling algorithms can be slow to
converge in high dimensional models~\cite{robert-04}.

Blei, Ng, and Jordan presented the first approximate inference technique for LDA
based on variational methods~\cite{blei-03}, but the collapsed
Gibbs sampler proposed by Griffiths and Steyvers~\cite{griffiths-04} has been
more popular in the community because it is easier to implement.  However, such
methods also have intrinsic problems that lead to difficulties in moving to
web-scale: a shared state, many short iterations, and randomness.

\paragraph{Shared State} 

Unless the probabilistic model allows for discrete segments to be statistically
independent of each other, it is difficult to conduct inference in
parallel. However, we want models that allow specialization to be shared across
many different corpora and documents when necessary, so we typically cannot
assume this independence.

At the risk of oversimplifying, collapsed Gibbs sampling for LDA is essentially
multiplying the number of occurrences of a topic in a document by the number of
times a word type appears in a topic across all documents.  The former is a
document-specific count, but the latter is shared across the entire corpus.  For
techniques that scale out collapsed Gibbs sampling for LDA, the major challenge
is keeping these second counts for collapsed Gibbs sampling consistent when
there is not a shared memory environment.

Newman et al.~\cite{newman-08} consider a variety of methods to achieve
consistent counts: creating hierarchical models to view each slice as
independent or simply syncing counts in a batch update.  Yan et
al.~\cite{yan-09} first cleverly partition the data using integer programming (an
NP-Hard problem).  Wang et al.~\cite{wang-09} use message passing to ensure that
different slices maintain consistent counts.  Smola and
Narayanamurthy~\cite{smola-10} use a distributed memory system to achieve
consistent counts.

Gibbs sampling approaches to scaling thus face a difficult dilemma: completely
synchronize counts, which compromises scaling, or allow for inconsistent counts,
which could negatively impact the quality of inference.  Many approaches take
the latter approach; sometimes the differences are negligible~\cite{newman-08},
but other times log-likelihood of the model trained by a single machines yields
a order of magnitude higher than the model trained by a cluster of
machines~\cite[Figure 4]{smola-10}.\footnote{They report log-likelihood for
  PubMed dataset is -0.8e+09 in single machine LDA but -0.7e+10 in multi-machine
  LDA; and log-likelihood for News dataset is -1.8e+09 in single machine LDA but
  -3.6e+10 in multi-machine LDA.}
%Many other approaches based on Gibbs sampling also suffered from this problem~\cite{wang-09, newman-08}.

In contrast to these engineering work-arounds, variational inference provides a
\emph{mathematical} solution of how to scale inference for LDA.  By assuming a
variational distribution that treats documents as independent, we can
parallelize inference without a need for synchronizing counts (as required in
collapsed Gibbs sampling).

\paragraph{Randomness}

By definition, Monte Carlo algorithms depend on randomness.  However, MapReduce
implementations assume that every step of computation will be the same, no
matter where or when it is run. This allows MapReduce to have greater
fault-tolerance, running multiple copies of computation steps in case a copy
fails or takes too long. Thus, MapReduce tasks cannot truly be random, which
against the nature of MCMC algorithms such as Gibbs sampling. This constraint
forces workarounds to ensure ``deterministic'' MCMC, for example seeding the
random number generator in a shard-dependent way~\cite{wang-09}.

\paragraph{Many short iterations}

A single iteration of Gibbs sampling for LDA with $K$ topics is very quick. For
each word, the algorithm performs a simple multiplication to build a sampling
distribution of length $K$, samples from that distribution, and updates an
integer vector. In contrast, each iteration of variational inference is
difficult; it requires the evaluation of complicated functions that are not
simple arithmetic operations directly implemented in an ALU (these are described
in Section~\ref{sec:model}).

This does not mean that variational inference is slower, however.  Variational
inference typically requires dozens of iterations to converge, while Gibbs
sampling requires thousands (determining convergence is often more difficult for
Gibbs sampling). Moreover, the requirement of Gibbs sampling to keep a
consistent state means that there are many more synchronizations required to
complete inference, increasing the complexity of the implementation and the
communication overhead. In contrast, variational inference requires
synchronization only once per iteration (dozens of times for a typical corpus);
in a na\"ive Gibbs sampling implementation, inference requires synchronization
after every word in every iteration (potentially billions of times for a
moderately-sized corpus).

\subsection{Variational Inference}

An alternative to MCMC is variational inference. Variational
methods, which are based on related techniques from statistical physics, use
optimization to find a distribution over the latent variables that is close to
the posterior of interest~\cite{jordan-99,wainwright-08}. Variational
methods provide effective approximations in topic models and nonparametric
Bayesian models~\cite{blei-05,teh-06,kurihara-07}.  We believe
that it is well-suited to MapReduce.

Variational methods enjoy clear convergence criterion, tend to be faster
than MCMC in high-dimensional problems, and provide particular
advantages over sampling when latent variable pairs are not conjugate.  Gibbs
sampling requires conjugacy, and other forms of sampling that can handle
non-conjugacy, such as Metropolis-Hastings, are much slower than variational
methods.

With a variational method, we begin by positing a family of distributions $q \in
Q$ over the same latent variables $Z$ with a simpler dependency pattern than
$p$, parameterized by $\Theta$. This simpler distribution is called the
variational distribution and is parameterized by $\Omega$, a set of variational
parameters.  With this variational family in hand, we optimize the
\textit{evidence lower bound} (ELBO),
\begin{equation}
  \elbo = \e{q}{\log \left(p(\mathbf{D} | Z) p(Z | \Theta) \right) } - \e{q}{\log q(Z)}
  \label{eqn:elbo}
\end{equation}
a lower bound on the data likelihood. Variational inference fits the variational
parameters $\Omega$ to tighten this lower bound and thus minimizes the
Kullback-Leibler divergence between the variational distribution and the
posterior.

The variational distribution is typically chosen by removing probabilistic
dependencies from the true distribution. This makes inference tractable
and also induces independence in the variational distribution between latent
variables. This independence can be engineered to allow paralleization of
independent components across multiple computers.

Maximizing the global parameters in MapReduce can be handled in a manner
analogous to EM~\cite{wolfe-08}; the expected counts (of the variational
distribution) generated in many parallel jobs are efficiently aggregated and
used to recompute the top-level parameters.

\subsection{Related Work}

Nallapati, Cohen and Lafferty~\cite{nallapati-07} extended variational inference
for LDA to a parallelized setting. Their implementation uses a master-slave
paradigm in a distributed environment, where all the slaves are responsible for
the E-step and the master node gathers all the intermediate outputs from the
slaves and performs the M-step. While this approach parallelizes the process to
a small-scale distributed environment, the final aggregation/merging showed an
I/O bottleneck that prevented scaling beyond a handful of slaves because the
master has to explicitly read all intermediate results from slaves.

Mr. LDA addresses these problems by parallelizing the work done by a single
master (a reducer is only responsible for a single topic) and relying on the
MapReduce framework, which can efficiently marshal communication between compute
nodes. Building on the MapReduce framework also provides advantages for
reliability and monitoring not available in an \emph{ad hoc} parallelization
framework.

% Distributing Gibbs sampling inference is also possible both in conventional
% parallelization frameworks~\cite{asuncion-09} and in graphics processing
% units~\cite{yan-09}.  Parallelizing Gibbs sampling requires either assuming
% strict independence between parallel components or maintaining complex
% synchronization techniques.  The former potentially hurts the accuracy of
% inference and the latter makes implementation and extension more difficult.

% Wang~\emph{et al.}~\cite{wang-09} extended the distributed Gibbs sampling
% framework to MapReduce, but still ignore the effect that partitioning can have
% on inference results.  Wang~\emph{et al.} also proposed a Message Passing
% Interface (MPI) which does not make these additional assumptions but which
% does require complicated data structures to ensure synchronization; we avoid
% that complication by using the MapReduce synchronization mechanisms.

% Moreover, as previously mentioned, Gibbs sampling requires more (but cheaper)
% iterations to converge.  Gibbs sampling is also an inherently stochastic
% process (as compared to variational inference, which is deterministic), but
% MapReduce tasks must be deterministic so that failed computations can be
% seamlessly replaced by taking over computation on a new node.

The MapReduce~\cite{dean-04} framework was originally inspired from the map and
reduce functions commonly used in functional programming. It adopts a
divide-and-conquer approach. Each \emph{mapper} processes a small subset of data
and passes the intermediate results as key value pairs to \emph{reducers}.  The
reducers recieve these inputs in sorted order, aggregate them, and produce the
final result. In addition to mappers and reducers, the MapReduce framework
allows for the definition of \emph{combiners} and \emph{partitioners}. Combiners
perform local aggregation on the key value pairs after map function. Combiners
help reduce the size of intermediate data transferred and are widely used to
optimize a MapReduce process. Partitioners control how messages are routed to
reducers.

%Thus it takes the same input as the reducer; by doing some of the work of the reducer in the mapper, it 

Mahout~\cite{mahout}, an open-source machine learning package, provides a
MapReduce implementation of variational inference LDA, but it lacks features
required by mature LDA implementations such as supplying per-document topic
distributions, computing likelihood, and optimizing hyperparameters (for an
explanation of why this is essential for model quality, see Wallach et al.'s
``Why Priors Matter''~\cite{wallach-09b}). Without likelihood computation, it's
impossible to know when inference is complete. Without per-document topic
distributions and likelihood bound estimates, it is impossible to quantitatively
compare performance with other implementations.

Table~\ref{tbl:comparison} provides a general overview and comparison of
features among different approaches for scaling LDA.

\begin{table*}
\scriptsize{
\center
\begin{tabular}{c|cccccccc}%p{1.2cm}p{1.2cm}p{1.2cm}p{1.2cm}}
\hline
%\multirow{2}{*}{Property} & \multicolumn{8}{|c}{Implementation} \\
%\cline{2-9}
& Mallet~\cite{mallet} & GPU-LDA~\cite{yan-09} & Async-LDA~\cite{asuncion-09} & N.C.L.~\cite{nallapati-07} & pLDA~\cite{wang-09} & Y!LDA~\cite{smola-10} & Mahout~\cite{mahout} & Mr. LDA\\
\hline
Framework & Multi-thread & GPU & Multi-thread &  Master-Slave & MPI \& MapReduce & Hadooop & MapReduce & MapReduce \\
\hline
Inference & Gibbs & Gibbs \& V.B. & Gibbs & V.B. & Gibbs & Gibbs & V.B. & V.B. \\
\hline
Likelihood Computation & Yes & Yes & Yes & N.A. & N.A. & Yes & No & Yes \\
\hline
Asymmetric $\alpha$ Prior & Yes & No & No & No & No & Yes & No & Yes\\
\hline
Hyperparameter Optimization & Yes & No & Yes & No & No & Yes & No & Yes \\
\hline
Informed $\beta$ Priors & No & No & No & No & No & No & No & Yes\\
\hline
Multilingual & Yes & No & No & No & No & No & No & Yes\\
\hline
\end{tabular}
}
\caption{Comparison among Different Approaches. (N.A. - not available from the paper context.)}
\label{tbl:comparison}
\end{table*}

\section{Mr. LDA}
\label{sec:model}

LDA assumes the following generative process to create a corpus of $M$ documents
with $N_d$ words in document $d$ using $K$ topics.
\begin{small}
\begin{enumerate*}
 \item For each topic index $k \in \{1, \dots K\}$, draw topic distribution
   ${\bm \beta_k} \sim \dir{{\bm \eta_k}}$
 \item For each document $d \in \{1, \dots M\}$:
   \begin{enumerate*}
   \item Draw document's topic distribution ${\bm \theta_{d}} \sim \dir{{\bm
       \alpha}}$
   \item For each word $n \in \{1, \dots N_d\}$:
    \begin{enumerate*}
     \item Choose topic assignment $z_{d,n} \sim \mult{{\bm \theta_d}}$
     \item Choose word $w_{d,n} \sim \mult{{\bm \beta_{z_{d,n}}}}$
     \end{enumerate*}
  \end{enumerate*}
\end{enumerate*}
\end{small}
In this process, $\dir{}$ represents a Dirichlet distribution, and $\mult{}$ is
a multinomial distribution. ${\bm \alpha}$ and ${\bm \beta}$ are parameters.

The mean-field variational distribution $q$ for LDA breaks the connection
between words and documents
\begin{equation*}
q(z, \theta, \beta) = \prod_k \dir{\beta_k \g \lambda_k} \prod_d \dir{\theta_d
  \g \gamma_d} \mult{z_{d,n} \g \phi_{d,n}},
\end{equation*}
which when used in Equation~\ref{eqn:elbo} can be used to derive updates that
optimize $\mathcal{L}$, the lower bound on the likelihood. In the sequel, we
take these updates as given, but interested readers can refer to the appendix of
Blei et al.~\cite{blei-03}.  Variational EM alternates between updating the
expectations of the variational distribution $q$ and maximizing the probability
of the parameters given the ``observed'' expected counts.

The remainder of the paper focuses on adapting these updates into the MapReduce
framework and challenges of working at a large scale.  We focus on
the primary components of a MapReduce algorithm: the mapper, which processes a
single unit of data (in this case, a document); the reducer, which processes a
single view of globally shared data (in this case, a topic parameter); the
partitioner, which allows order inversion for normalization; and the driver,
which controls the overall algorithm.  The interconnections between the
components of Mr. LDA are depicted in Figure~\ref{fig:workflow}.

\subsection{Mapper: Update $\phi$ and $\gamma$}
\label{sec:model:mapper}

Each document has associated variational parameters $\gamma$ and $\phi$.  The
mapper computes the updates for these variational parameters and uses them to
create the sufficient statistics needed to update the global parameters.  In
this section, we describe the computation of these variational updates and how
they are transmitted to the reducers.

Given a document, the updates for $\phi$ and $\gamma$ are
\begin{equation*}
\phi_{v, k} \propto \e{q}{\beta_{v, k}} \cdot e^{\digam{\gamma_{k}}},
\phantom{spac} \gamma_{k} = \alpha_{k} + \sum_{v=1}^{V} \phi_{v, k},
\end{equation*}
where $v \in [1, V]$ is the term index and $k \in [1, K]$ is the topic index. In
this case, $V$ is the size of the vocabulary $\mathcal{V}$ and $K$ denotes the
total number of topics.  The expectation of $\beta$ under $q$ gives an estimate
of how compatible a word is with a topic; words highly compatible with a topic
will have a larger expected $\beta$ and thus higher values of $\phi$ for that
topic.

Algorithm~\ref{alg:mapper} illustrates the detailed procedure of the Map
function. In the first iteration, mappers initialize variables, e.g. seed
$\lambda$ with the counts of a single document. For the sake of brevity, we omit
that step here; in later iterations, global parameters are stored in
\emph{distributed cache} -- a synchronized read-only memory that is shared
among all mappers~\cite{white-10} -- and retrieved prior to mapper execution.

A document is represented as a term frequency sequence $\vec{w} = \Vert w_1,
w_2, \ldots, w_V \Vert$, where $w_i$ is the corresponding \textbf{term 
frequency in document $d$}. For ease of notation, we assume the input term
frequency vector $\vec{w}$ is associated with all the terms in the vocabulary,
i.e., if term $t_i$ does not appear at all in document $d$, $w_i=0$.

Because the document variational parameter $\gamma$ and the word variational
parameter $\phi$ are tightly coupled, we impose a local
convergence requirement on $\gamma$ in the Map function.  This means that the
mapper alternates between updating $\gamma$ and $\phi$ until $\gamma$ stops
changing.

\begin{algorithm}
\begin{small}
    \caption{Mapper}
    \label{alg:mapper}
%    \begin{footnotesize}
	\vspace{1ex}\noindent\textbf{Input:}\\
	\textsc{Key} - document ID $d \in [1, C]$, where $C = \vert \mathcal{C} \vert$.\\
	%is the total number of documents, i.e., size of the collection.\\
	\textsc{Value} - document content.
		
%	\vspace{1ex}\noindent\textbf{Output:}\\
%		\textsc{Key} - key pair $\langle p_l, p_r \rangle$.\\
%		\textsc{Value} - value $\sigma^\prime$.
	
% 	\vspace{1ex}\noindent\textbf{Configuration}
% 	\begin{algorithmic}[1]
% %		\STATE Initialize the total number of topics as $K$, size of the vocabulary $\mathcal{V}$ as $V$, size of the collection $\mathcal{C}$ as $C$.
% 		\STATE Initialize all the necessary parameter $K$, $V$ and $C$.
% 		\STATE Initialize or seed a $V \times K$ dimensional matrix $\beta$.
% 		\STATE Initialize or seed a $C \times K$ dimensional matrix $\gamma$.
% 		\STATE Initialize or seed a $K$ dimensional vector $\alpha$.
% 	\end{algorithmic}
	
	\vspace{1ex}\noindent\textbf{Map}
%        \end{footnotesize}

%        \begin{footnotesize}
	\begin{algorithmic}[1]
	  \STATE Initialize a zero $V \times K$-dimensional matrix $\phi$.
	  \STATE Initialize a zero $K$-dimensional row vector $\sigma$.
	  \STATE Read in document content $\Vert w_1, w_2, \ldots, w_V \Vert$
	  \REPEAT
	  \FORALL{$v \in [1, V]$}
          \label{alg:update-phi-start}
	  \FORALL{$k \in [1, K]$}
	  \STATE Update $\phi_{v, k}= \frac{\lambda_{v, k}}{\sum_v{\lambda_{v,k}}} \cdot \exp(\digam{\gamma_{d, k}})$.
	  \ENDFOR
	  \STATE Normalize $\phi_{v}$, set $\sigma = \sigma + w_v \phi_{v, *}$
	  \ENDFOR
	  \STATE Update row vector $\gamma_{d, *} = \alpha + \sigma$.
          \label{alg:update-phi-end}
	  \UNTIL{convergence}
	  \FORALL{$k \in [1, K]$}
	  \FORALL{$v \in [1, V]$}
	  \STATE Emit $\langle k, \triangle \rangle : w_v \phi_{v, k}$.
	  \COMMENT{Section~\ref{sec:model:partitioner}}
	  \label{alg:emit-marginal}
	  \STATE Emit $\langle k, v \rangle : w_v \phi_{v, k}$.
	  \COMMENT{order inversion}
	  \ENDFOR
	  \STATE Emit $\langle \triangle, k \rangle : (\digam{\gamma_{d, k}} - \digam{\sum_{l=1}^K \gamma_{d,l}})$.
	  \COMMENT{$\alpha$ update, Section~\ref{sec:model:driver}}
	  \STATE Emit $\langle k, d \rangle - \gamma_{d, k}$ to file.
	  \ENDFOR
	  \STATE Emit $\langle \triangle, \triangle \rangle -
          \mathcal{L}$ \COMMENT{ELBO,
            Section~\ref{sec:model:likelihood}}
	\end{algorithmic}
%        \end{footnotesize}
\end{small}
\end{algorithm}

\subsection{Partitioner: Efficient Marginal Sums}
\label{sec:model:partitioner}

The Map function in Algorithm~\ref{alg:mapper} emits sufficient statistics,
which we will need to compile and normalize. To take advantage of the MapReduce
framework to handle this computation, we use the \emph{order inversion} design
pattern~\cite{lin-10,lin-09}.

\begin{figure}[htb]
\begin{center}
\subfigure[LDA]{
  \label{fig:graphmod:lda}
  \includegraphics[width=0.4\linewidth]{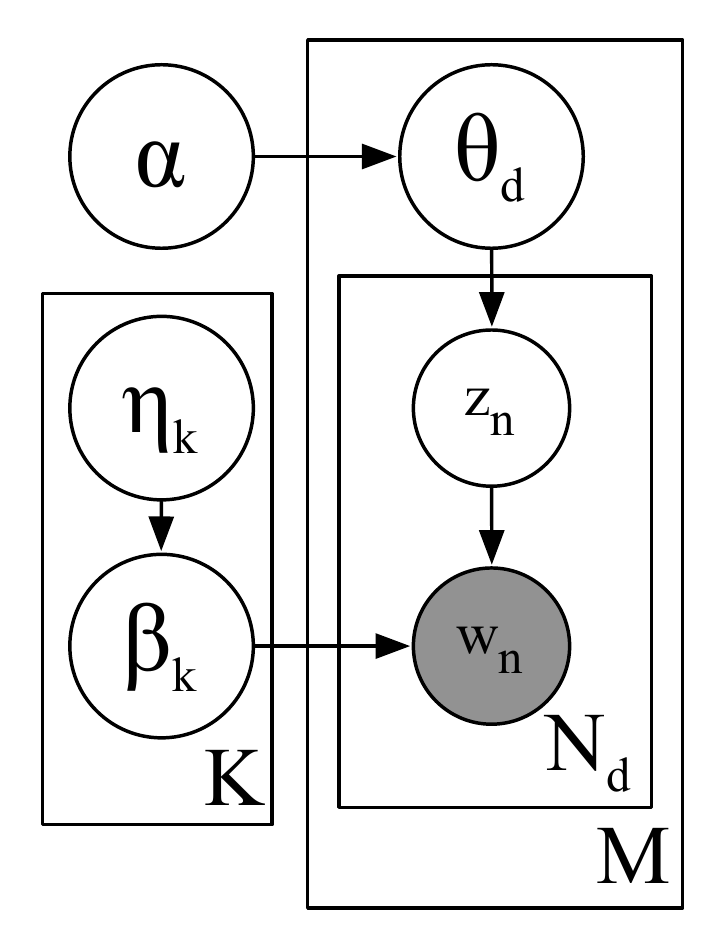}
}
\subfigure[Variational]{
  \label{fig:graphmod:variational}
  \includegraphics[width=0.4\linewidth]{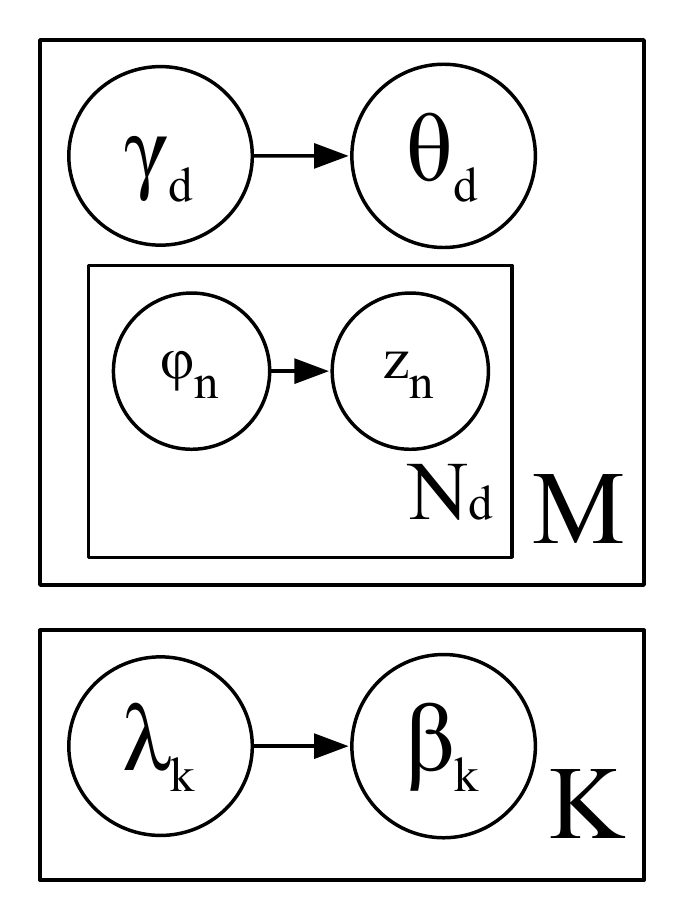}
}

\caption{Graphical model of LDA and the mean field variational distribution.
  Each latent variable, observed datum, and parameter is a node.  Lines
  between represent possible statistical dependence.  Shaded nodes
  are observations; rectangular plates denote replication; and numbers in
  the bottom right of a plate show how many times plates' contents repeat.
  In the variational distribution (Figure~\ref{fig:graphmod:variational}), the
  latent variables $\theta$, $\beta$, and $z$ are explained by a simpler, fully
  factorized distribution with variational parameters $\gamma$, $\lambda$, and
  $\phi$.  The lack of inter-document dependencies in the variational
  distribution allows the parallelization of inference in the MapReduce.}
\label{fig:graphmod}
\end{center}
\end{figure}

The sufficient statistics are keyed by a composite key set $\langle
p_{\mathsf{left}}, p_{\mathsf{right}} \rangle$. It is normally a pair of topic
and word identifier.  There are three exceptions:
\begin{small}
\begin{itemize}
\item In addition to the vocabulary terms, we choose a special normalization key
  value (denoted by $\triangle$) that comes before any ``normal'' vocabulary
  key, i.e., $\triangle < v, \forall v \in [1, V]$.  Before any word tokens are
  seen by the reducer, the reducer can compute the normalization term -- by
  summing all of the values associated with $\triangle$ -- and afterward write
  the final parameter values in a single pass through the reducer's keys.
\item If the value represents the sufficient statistics for $\alpha$ updating, the key pair is $\triangle$ and a
  topic identifier.  The MapReduce framework ensures those keys arrive in
  lexicographic sorted order.

\item Finally, if both keys are $\triangle$, it represents a document's
  contribution to the likelihood bound $\elbo$; this is combined with the
  topic's contribution.
\end{itemize}
\end{small}

This assumes that, for calculating a new parameter, a single reducer will see
both the normalization key and all word keys. This is accomplished by ensuring
the partitioner sorts on topic only. Thus, any reducers beyond the number of
topics is superfluous.  Given that the vast majority of the work is in the
mappers, this is typically not an issue for LDA.

%Algorithm~\ref{alg:partitioner} illustrates the pseudo-code of the partitioner.

%\begin{algorithm}
%	\caption{Partitioner}
%	\label{alg:partitioner}
%
%	\begin{scriptsize} 
%	\vspace{1ex}\noindent\textbf{Input:}\\
%		\textsc{Key} - key pair $\langle p_l, p_r \rangle$.
%		
%	\vspace{1ex}\noindent\textbf{Partitioner}
%	\begin{algorithmic}[1]
%		\STATE Partition data according to $p_l$ only, ignore $p_r$.
%	\end{algorithmic}
%	\end{scriptsize}
%\end{algorithm}

%\jbgcomment{Didn't understand this paragraph - should we cut:}

%Order inversion is a common implementation routine in MapReduce framework to
%compute a probability distribution in an efficient way. In learning and 
%inferencing a non-parametric Bayesian model, this programming pattern 
%could be widely used for computing the likelihood as well as the posterior distribution.

\subsection{Reducer: Update $\lambda$}
\label{sec:model:reducer}

The Reduce function updates the variational parameter $\lambda$ associated with each topic.  Because of the order inversion described above, the update is
straightforward. It requires aggregation over all intermediate $\phi$ vectors
\begin{align*}
\lambda_{v, k} & = \eta_{v,k} + \sum_{d=1}^{C} ( w^{(d)}_v \phi^{(d)}_{v, k}),
\end{align*}
where $d \in [1, C]$ is the document index and $w^{(d)}_v$ denotes the number of
appearances of term $v$ in document $d$. Similarly, $C$ is the number of
documents.\footnote{Although the variational update for $\lambda$ does not
  include a normalization, the expectation $\e{q}{\beta}$ requires the $\lambda$
  normalizer.  In practice, the value $\frac{\lambda_{v,k}}{\sum_{w}
    \lambda_{w,k}}$ is distributed to mappers in the next iteration.}

This is elaborated in Algorithm~\ref{alg:reducer}; Step 10 completes the final
procedure for the order inversion design pattern -- collecting all the marginal
distribution counts.  These aggregations will then be written to parameter files
that will be used in subsequent iterations.  Step 15 adds the contribution of the
topic to the overall likelihood.

\begin{algorithm}
\begin{small}
    \caption{Reducer}
    \label{alg:reducer}
    \vspace{1ex}\noindent\textbf{Input:}\\
    \textsc{Key} - key pair $\langle p_{\mathsf{left}}, p_{\mathsf{right}} \rangle$.\\
    \textsc{Value} - an iterator $\mathcal{I}$ over sequence of values.
		
%	\vspace{1ex}\noindent\textbf{Output:}\\
%		\textsc{Key} - topic index $k \in [1, K]$.\\
%		\textsc{Value} - $V$-dimensional column vector $\phi_{*, k}$.
\hidetext{
    \vspace{1ex}\noindent\textbf{Configuration}
    \begin{algorithmic}[1]
          \STATE Initialize normalizer $\nu= 0$, ELBO $\tau = 0$
    \end{algorithmic}
}
	\vspace{1ex}\noindent\textbf{Reduce}
	\begin{algorithmic}[1]
          \STATE Compute the sum $\sigma$ over all values in the sequence
          $\mathcal{I}$.

          \IF{$p_{\mathsf{left}}=\triangle$} \IF{$p_{\mathsf{right}}=\triangle$}
             \STATE $\tau = \tau + \sigma$
             \COMMENT{ELBO $\mathcal{L}$}
         \ELSE
         \STATE Emit $\langle \triangle, p_{\mathsf{right}} \rangle : \sigma$
          \COMMENT{$\alpha$ update, Section~\ref{sec:model:driver}}
         \ENDIF
          \ELSE 
          \IF{$p_{\mathsf{right}}=\triangle$} 
          \STATE Normalizer $\nu = \sigma + \sum_{v} \eta_{v,k}$.  \COMMENT{order inversion}
          \STATE $\tau = \tau - \LG{\nu}$
          \label{reducer step: collect marginal distribution}
          \ELSE
          \STATE $\lambda_{v,k} = \eta_{v,k} + \sigma$
          \STATE Emit $\langle k, v \rangle : \frac{\lambda_{v,k}}{\nu + \sum_{j} \eta_{j,k}}$. 
          \COMMENT{normalized $\e{q}{\beta}$ value}
          \STATE $\tau = \tau + \LG{\lambda_{v,k}} + \left(\lambda_{v,k} - 1 \right) \left( \digam{\lambda_{v,k}} - \digam{\nu}  \right)$
			\ENDIF
		\ENDIF
          \STATE Emit $\langle \triangle, \triangle \rangle : \tau$
	\end{algorithmic}
\end{small}
\end{algorithm}

To improve performance, we use combiners to facilitate the aggregation of sufficient
statistics in mappers before they were transferred to reducers.  This decreases
bandwidth and saves the reducer computation.

%\begin{algorithm}
%	\label{alg:combiner}
%	\caption{Combiner}
%	
%	\begin{scriptsize}
%	
%	\vspace{1ex}\noindent\textbf{Input:}\\
%		\textsc{Key} - $\langle$topic, term$\rangle$ index pair $\langle k, v \rangle, k \in [1, K], v \in [1, V]$.\\
%		\textsc{Value} - an iterator $\mathcal{I}$ over sequence of $V$-dimensional column vector $\phi_{*, k}$.
%		
%	\vspace{1ex}\noindent\textbf{Output:}\\
%		\textsc{Key} - topic index $k \in [1, K]$.\\
%		\textsc{Value} - $V$-dimensional column vector $\beta$.
%	
%	\vspace{1ex}\noindent\textbf{Reduce}
%	\begin{algorithmic}[1]
%		\STATE Initialize a $V$-dimensional column vector $\beta$.
%		\WHILE{$\mathcal{I}$ has next element}
%			\STATE Denote $\mathcal{I}$'s next element is $\phi_{*, k}$.
%			\STATE Update $\beta=\beta+\phi_{*, k}$.
%		\ENDWHILE
%		\STATE Output key-value pair $k - \phi_{d, k}$ to file.
%	\end{algorithmic}
%	\end{scriptsize}
%\end{algorithm}

\begin{figure}[htb]
\begin{center}
 \includegraphics[width=1.0\linewidth]{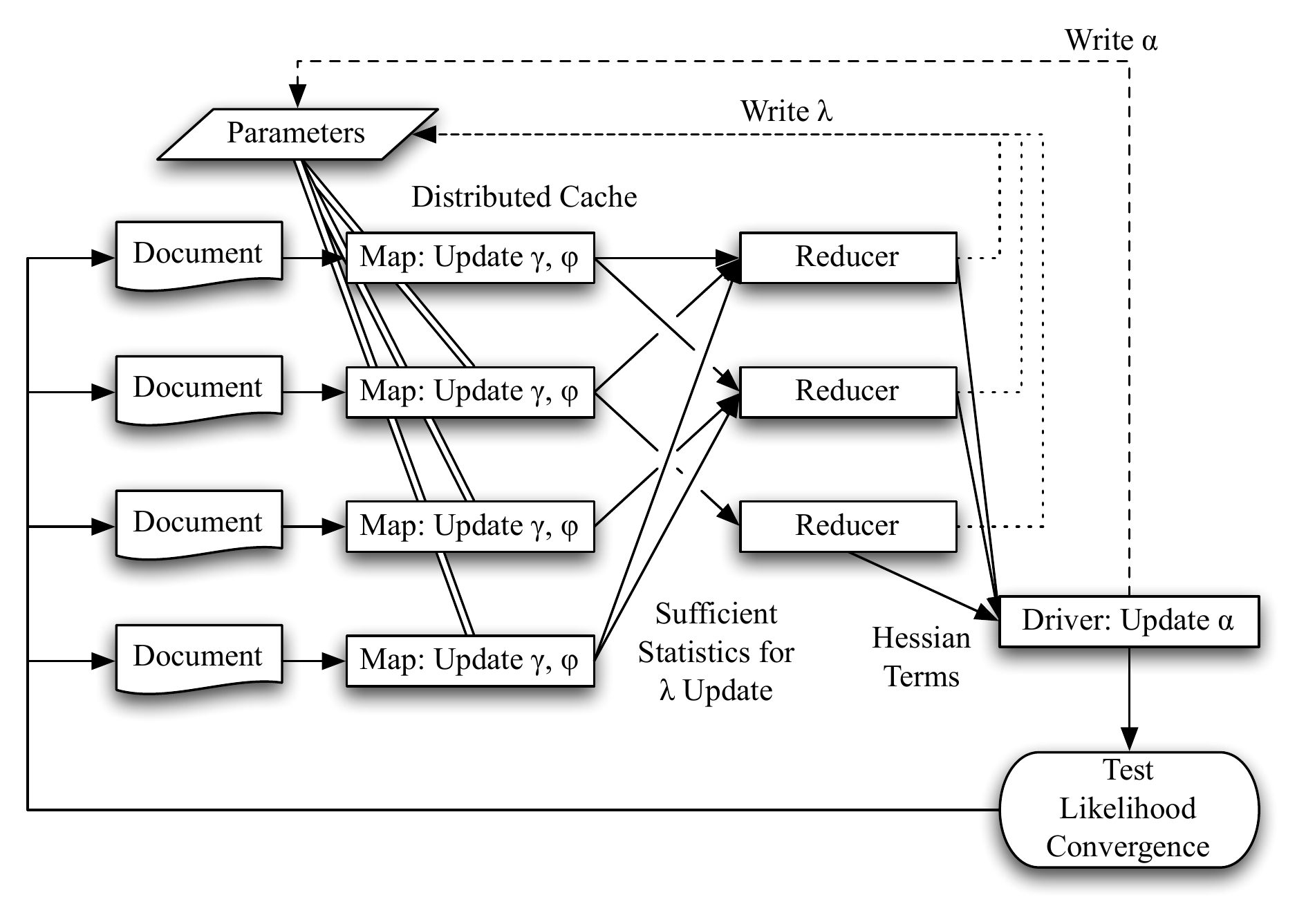}
 \caption{Workflow of Mr. LDA.  Each iteration is broken into three stages:
   computing document-specific variational parameters in parallel mappers,
   computing topic-specific parameters in parallel reducers, and then updating
   global parameters in the driver, which also monitors convergence of the
   algorithm.  Data flow is managed by the MapReduce framework: sufficient
   statistics from the mappers are directed to appropriate reducers, and new
   parameters computed in reducers are distributed to other computation units
   via the distributed cache.}
\jbgcomment{Update image to include $\lambda$}
\label{fig:workflow}
\end{center}
\end{figure}

\subsection{Driver: Update $\alpha$}
\label{sec:model:driver}

Effective inference of topic models depends on learning not just the latent
variables $\beta$, $\theta$, and $z$ but also estimating the hyperparameters,
particularly $\alpha$.  The $\alpha$ parameter controls the sparsity of topics
in the document distribution and is the primary mechanism that differentiates
LDA from previous models like pLSA and LSA; not optimizing $\alpha$ risks
learning suboptimal topics~\cite{wallach-09b}.

Updating hyperparameters is also important from the perspective of equalizing
differences between inference techniques; as long as hyperparameters are
optimized, there is little difference between the \emph{output} of inference
techniques~\cite{asuncion-09b}.

The driver program marshals the entire inference process.  On the first
iteration, the driver is responsible for initializing all the model parameters
($K$, $V$, $C$, $\eta$, $\alpha$); the number of topics $K$ is user specified;
$C$ and $V$, the number of documents and types, is determined by the data; the
initial value of $\alpha$ is specified by the user; and $\lambda$ is randomly
initialized or seeded by documents.

\subsection{Likelihood Computation}
\label{sec:model:likelihood}

The driver monitors the ELBO to determine whether inference has converged. If
not, it restarts the process with another round of mappers and reducers.
To compute the ELBO we expand Equation~\ref{eqn:elbo}, which gives us
\begin{footnotesize}
\begin{align*}
\elbo(\gamma, \phi, \lambda; \alpha,
\eta)= & \explain{\footnotesize{driver}}{\sum_{d=1}^C \Phi(\alpha)}
  + \explain{\footnotesize{computed in reducer}}{\sum_{d=1}^C 
        (\explain{\footnotesize{computed in mapper}}{\mathcal{L}_d(\gamma, \phi)  + 
          \mathcal{L}_d(\phi) - \Phi(\gamma)})} \\
& + \explain{\footnotesize{driver / constant}}{ \sum_{k=1}^{K} \Phi(\eta_{*,k})} - \explain{\footnotesize{driver}}{ \sum_{k=1}^{K} \explain{\footnotesize{reducer}}{\Phi(\lambda_{*,k}) }    }
\end{align*}
\end{footnotesize}
where
\begin{small}
\begin{align*}
\Phi(\mu) = & \LG{\sum_{i=1} \mu_i} - \sum_{i=1} \LG{\mu_i} \\
    & + \sum_i \left(\mu_i - 1 \right) \left( \digam{\mu_i} - \digam{\sum_j \mu_j} \right). \\
\mathcal{L}_d(\gamma, \phi)  = & \sum_{k=1}^K \sum_{v=1}^V \phi_{v, k} w_v \left[
\digam{\gamma_k}-\digam{\sum_{i=1}^K \gamma_i} \right],\\
\mathcal{L}_d(\phi)  = & \sum_{v=1}^V \sum_{k=1}^K \phi_{v, k} \left(\sum_{i=1}^V w_i \log \frac{\lambda_{i, k}}{\sum_j \lambda_{j, k}} - \log \phi_{v, k} \right),
\end{align*}
\end{small}
Almost all of the terms that appear in the likelihood term can be computed in
 mappers; the only term that cannot are the terms that depend on $\alpha$,
which is updated in the driver, and the variational parameter $\lambda$, which
is shared among all documents. All terms that depend on $\alpha$ can be easily
computed in the driver, while the terms that depend on $\lambda$ can be computed
in each reducer.

Thus, computing the total likelihood proceeds as follows: each mapper computes
its contribution to the likelihood bound $\elbo$, and emits a special key that
is unique to likelihood bound terms and then aggregated in the reducer; the
reducers add topic-specific terms to the likelihood; these final values are then
combined with the contribution from $\alpha$ in the driver to compute a final
likelihood bound.

\hidetext{

\begin{algorithm}
  \begin{scriptsize}
    \caption{Driver}
    \label{alg:driver}	
	\vspace{1ex}\noindent\textbf{Input:}\\
		$K$ - total number of topics\\
		$V$ - total size of the vocabulary\\
		$C$ - total number of documents in the collection
%		$\alpha$ - a $K$ dimensional vector with non-zero initial value.\\
%		$\beta$ - a $V \times K$ dimensional matrix, either random or seeded, subject to $\displaystyle\sum_{v=1}^V\beta_{v, *}=1$
	
	\vspace{1ex}\noindent\textbf{Driver}
	\begin{algorithmic}[1]
		\STATE Initialize a $K$ dimensional vector $\alpha$ with value $\alpha_0$.
		\STATE Initialize a $V \times K$ dimensional matrix $\lambda$
                (random or seeded).
		\REPEAT
			\STATE Distribute model parameters $K$, $V$, $C$ to the cluster.
            \STATE Distribute $\lambda$ and $\alpha$ to the cluster via \emph{distributed cache}.
            \STATE Mappers compute updates to $\gamma$, $\phi$.
			\STATE Reducers compute new $\lambda$.
			\STATE Read in all the $\gamma$-tokens ($\alpha$ sufficient statistics) from the output files.
			\STATE Compute global contribution to likelihood bound $\mathcal{L}$
			\REPEAT
				\STATE Update $\alpha$ according to Newton-Raphson method.
				%\STATE Compute likelihood function $L(\gamma, \phi; \alpha, \beta)$.
			\UNTIL{convergence}
		\UNTIL{convergence}
	\end{algorithmic}
  \end{scriptsize}
\end{algorithm}
}

The driver updates $\alpha$ after each MapReduce iteration. We use a
Newton-Raphson method which requires the Hessian matrix and the gradient,
\begin{align*}
\alpha_{\mathsf{new}} = \alpha_{\mathsf{old}} - \mathcal{H}^{-1}(\alpha_{\mathsf{old}}) \cdot g(\alpha_{\mathsf{old}}),
\end{align*}
where the Hessian matrix $\mathcal{H}$ and $\alpha$ gradient are defined respectively as
\begin{align*}
 \mathcal{H}(k, l) = & \delta(k, l) C \ddigam{\alpha_k} - C \ddigam{\sum_{l=1}^K \alpha_l},\\
 g(k) = & \explain{\footnotesize{computed in driver}}{C \left( \digam{ \sum_{l=1}^K \alpha_l} -
  \digam{\alpha_k} \right) } + \\
& \explain{\footnotesize{computed in reducer}}{\sum_{d=1}^C
  \explain{\footnotesize{computed in mapper}}{\digam{\gamma_{d, k}} - \digam{\sum_{l=1}^K
      \gamma_{d, l}}} }.
\end{align*}
The Hessian matrix $\mathcal{H}$ depends entirely on the vector $\alpha$, which
changes during updating $\alpha$. The gradient $g$, on the other hand, can be
decomposed into two terms: the \emph{$\alpha$-tokens} (i.e.,
$\digam{\sum_{l=1}^K \alpha_l} - \digam{\alpha_k}$) and the $\gamma$-tokens
(i.e., $\sum_{d=1}^C \digam{\gamma_{d, k}} - \digam{\sum_{l=1}^K \gamma_{d,
    l}}$).  We can remove dependence on the number of documents in the
gradient computation by computing the $\gamma$-tokens in mappers.  This key
observation allows us to optimize $\alpha$ in the MapReduce environment.

Because LDA is a dimensionality reduction algorithm, there are typically a small
number of topics $K$ even for a large document collection. As a result, we can
safely assume the dimensionality of $\alpha$, $\mathcal{H}$, and $g$ are
reasonably low, and additional gains come from the diagonal structure of the
Hessian~\cite{minka-00}. Hence, the updating of $\alpha$ is
efficient and will not create a bottleneck in the driver.

\section{Flexibility of Mr. LDA}
\label{sec:flexibility}

In this section, we highlight the flexibility of Mr. LDA to accomodate
extensions to LDA. These extensions are possible because of the modular nature
of Mr. LDA's design.

\subsection{Informed Prior}
\label{sec:flexibility:informed prior}

The standard practice in topic modeling is to use a same symmetric prior
(i.e. $\eta_{v,k}$ is the same for all topics $k$ and words $v$).  However, the
model and inference presented in Section~\ref{sec:model} allows for topics to
have different priors; allowing users to incorporate prior information into the
model.  

For example, suppose our we wanted to discover how different psychological
states were expressed in blogs or newspapers. If this were our goal, we might
reasonably create priors that captured psychological categories to discover how
they were expressed in a corpus. The Linguistic Inquiry and Word Count (LIWC)
dictionary~\cite{pennebaker-99} defines 68 categories encompassing psychological
constructs and personal concerns. For example, the \emph{anger} LIWC category
includes the words ``abuse,'' ``jerk,'' and ``jealous;'' the \emph{anxiety}
category includes ``afraid,'' ``alarm,'' and ``avoid;'' and the \emph{negative
  emotions} category includes ``abandon,'' ``maddening,'' and ``sob.''  Using
this dictionary, we built a prior ${\bm \eta}$ as follows:
\begin{equation*}
\eta_{v,k} =
\begin{cases}
  10, \text{ if } v \in \mbox{LIWC category}_k\\
0.01, \text{ otherwise}
\end{cases},
\end{equation*}
where $\eta_{v, k}$ is the informed prior for word $v$ of topic $k$. This is
accomplished via a slight modification of the reducer and leaving the rest of
the system unchanged.

\subsection{Polylingual LDA}
\label{sec:flexibility:polylingual}

In this section, we demonstrate the flexibility of Mr. LDA by showing how its
component-based design allows for extending LDA beyond a single language.
PolyLDA~\cite{mimno-09} assumes a document-aligned
multilingual corpus. For example, articles in Wikipedia have links to the version
of the article in other languages; while the linked documents are ostensibly on
the same subject, they are usually not direct translations, and possibly written to have a
culture-specific focus.

\begin{figure}
\begin{center}
\includegraphics[width=0.8\linewidth]{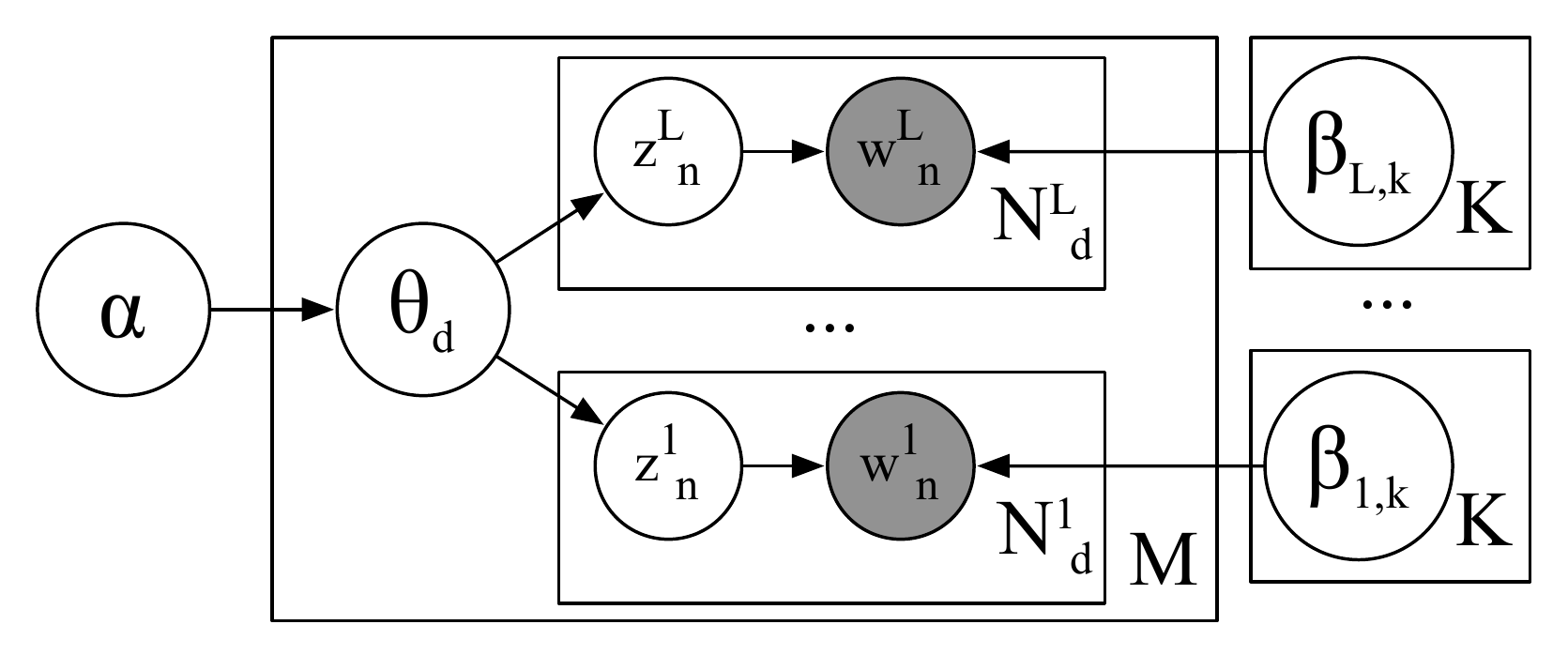}
\end{center}
\caption{Graphical model for polylingual LDA~\cite{mimno-09}.  Each document has
  words in multiple languages.  Inference learns the common topic ids across
  languages that co-occur in the corpus.  The modular inference of Mr. LDA
  allows for inference for this model to be accomplished by the same framework
  created for monolingual LDA.}
\label{fig:polylda}
\end{figure}

PolyLDA assumes that a single document has words in multiple languages, but each
document has a common, language agnostic per-document distribution $\theta$
(Figure~\ref{fig:polylda}).  Each topic also has different facets for language;
these topics end up being consistent because of the links across language
encoded in the consistent themes present in documents.

Because of the modular way in which we implemented inference, we can perform
multilingual inference by embellishing each data unit with a language identifier
$l$ and change inference as follows:
\begin{small}
\begin{itemize}
\item Updating $\lambda$ happens $l$ times, one for each language. The updates
  for a particular language ignores expected counts of all other languages.
\item Updating $\phi$ happens using only the relevant language for a word.
\item Updating $\gamma$ happens as usual, combining the contributions of all
  languages relevant for a document.
\end{itemize}
\end{small}

From an implementation perspective, PolyLDA is a collection of monolingual
Mr. LDA computations sequenced appropriately. Mr. LDA's approach of taking
relatively simple computation units, allowing them to scale, and preserving
simple communication between computation units stands in contrast to the design
choices by approaches using Gibbs sampling.

For example, Smola and Narayanamurthy~\cite{smola-10} interleave the topic and
document counts during the computation of the conditional distribution using Yao
et al.'s ``binning'' approach~\cite{yao-09}.  While this improves performance,
changing any of the modeling assumptions would potentially break this
optimization.

In contrast, Mr. LDA's philosophy allows for easier development of extensions of
LDA.  While we only discuss two extensions here, other extensions are
possible. For example, implementing supervised LDA~\cite{blei-07b} only requires
changing the computation of $\phi$ and a regression; the rest of the model is
unchanged. Implementing syntactic topic models~\cite{boyd-graber-08} requires
changing the mapper to incorporate syntactic dependencies.

\reviewercomment{The paper does not satisfactorily establish its novelty. As noted in
Section 5, there have been many other approaches for parallelizing
LDA. The authors dismiss those other approaches in various ways, yet
offer no empirical results comparing them nor showing any contrasting
tradeoffs in concrete detail. They also mention that Mahout has an LDA
implementation, but say it "lacks features required by a mature LDA
implementation". Are the authors arguing that the Mahout code needs to
be more "feature complete", or are they arguing that this proposed
approach provides ways to do this that cannot be straight-forwardly
done within Mahout?}

\hidetext{
\reviewercomment{There are several concerns about the technical quality of the paper as
written, both in the reasonableness of any claims and the empirical
support for them.}

\reviewercomment{This paper propose Mr.LDA, a parallelized LDA
    algorithm in MapReduce programming framework. I really appreciate that the
    author has implemented Mr. LDA on a cluster contains 900 nodes and evaluated
    its performance on a real-world large-scale corpus. However, because
    variational inference and distributed implementation for LDA are not new,
    the paper should show incremental and important contributions compared to
    previous work. Theoretically these is no contribution as update equations
    are given from the original LDA paper. From the engineering perspective, as
    this paper claims previous work must make compromises during inference and
    modeling, or previous methods have difficulties in moving to large scale,
    the advantages of the proposed method must be shown in the experiments.}  }

\reviewercomment{EXPERIMENTS:
The advantages of the proposed method must be shown in the
experiments. For example, how about the running efficiency and how
about the likelihood and perplexity compared to previous "compromised"
methods such as parallelized Gibbs LDA? It would be much better to
compare the performance with that of other existing implementations
such as Mahout, if possible. The experiment results in this paper are
only analyses of the proposed method, which are not exciting and not
convincing enough. To substantiate its contribution, this paper should
add comparative experimental results, and it would be a significant
plus if the author can publish the source code to help other
researchers use this framework.}

\section{Experiments}
\label{sec:exp}

We implemented Mr. LDA\footnote{Code available after blind review.} using Java
with Hadoop 0.20.1 and ran it on a cluster provided by NSF's \textbf{CLU}ster
\textbf{E}xploratory Program (CluE) and the Google/IBM Academic Cloud Computing
Initiative. The cluster used in our experiments contained 280 physical nodes;
each node has two single-core processors (2.8 GHz), 4 GB memory, and two 400 GB
hard drives. The cluster was configured to run a maximum of three map tasks and
two reduce tasks simultaneously, and usually under a heavy, heterogeneous load.

%--------------------------------------------------------
\hidetext{
\subsection{Scalability}

We first demonstrate that Mr. LDA scales well by showing the total computation
time as the number of documents increases. We can add more mappers to
effectively absorb the additional load.

We also show that we scale well as the number of topics increase.

We now show that Mr. LDA has better generalization and speed than alternative
methods that do not optimize hyperparameters (e.g. against Mahout).

\jbgcomment{In an ideal world, we would also compare against the other MapReduce
  LDA implementations in terms of speed.  I'm not sure we'd come out favorably
  at this point.  We should work on it, though.}

\jbgcomment{Another idea to make the case about iterations and time is to have a
  comparison with Mallet.  Give Mallet the same number of documents, see how
  long it takes for Mallet to converge(both in terms of number of iterations and
  total time) and the time per iteration.}

This gives users flexibility in the size of the dataset, but not necessarily
flexibility in terms of the kinds of the analysis they can do.

\subsection{Flexibility for Directed Exploration of Corpora}

\jbgcomment{These sections seem redundant given the (unexpanded) first
  subsection in this section - merge or delete one.  Make them all fit together
  in a way that flows from one section to the next.}

}
%--------------------------------------------------------

\subsection{Scalability}
\label{sec:exp:scaling}

\reviewercomment{Section 4.2 claims training time increases "sublinearly" with
  the number of input documents. That seems a strange claim -- the number of
  input bytes being processed (by the same fixed number of cluster nodes) should
  grow linearly with the documents, not sublinearly. It would seem that the only
  way sublinear growth could happen in practice is if the training times for
  some subset of the data is excessively high, meaning that the constants in the
  overhead complexity are high.}

We report results on the TREC document collection (disks 4 and 5
\cite{trec-22}), consisting mostly of newswire documents from the
\emph{Financial Times} and \emph{LA Times}. It contains more than $100,000$
distinct word types in approximately half a million documents. As a
preprocessing step, we remove types that appear fewer than $20$ times and apply
stemming~\cite{snowball}, reducing the vocabulary size to approximately
$65,000$.  This speeds inference and is consistent with standard approaches for
LDA (but with a larger vocabulary than is typical).

\begin{figure}[htb]
\begin{center}
\includegraphics[width=0.9\linewidth]{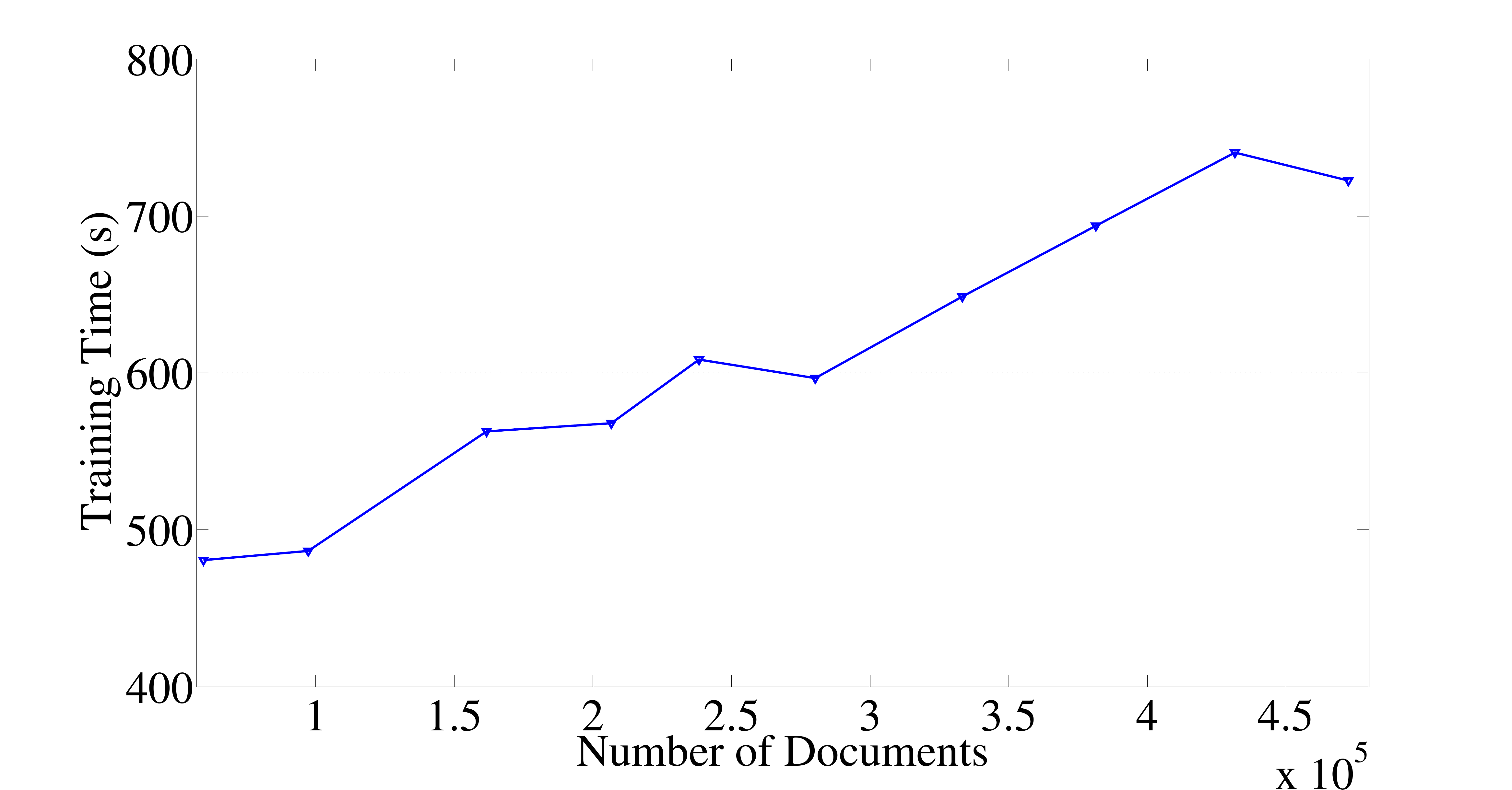}
\caption{Scalability vs. No. of Input Documents. The average training time
  increases in an approximately linear fashion, as the number of input
  documents increases. This suggests Mr. LDA is parallelizing effectively as more
  computing resources are available.}
\label{fig:training time scaling doc}
\end{center}
\end{figure}

Figure~\ref{fig:training time scaling doc} shows the relationship between
training time and corpus size the training time averaged over the first 20
Map/Reduce iterations. For this experiment, the number of topics was set to
$K=10$, and inference was done with 137 mappers (the number of input sequence
files) and 100 reducers\footnote{The number of reducers \emph{actually} used is
  limited by the number of topics because of the partitioning. However, later
  experiments, all $100$ reducers will be used, so the number of reducers is set
  to $100$ for consistency}.  Doubling the corpus size results in a less than
20\% increase increase in running time, suggesting that Mr. LDA is able to
successfully distribute the workload to more machines and take advantage of
parallelism. As the number of input documents increases, the training time
increases gracefully.

\begin{figure}[htb]
\begin{center}
\includegraphics[width=0.85\linewidth]{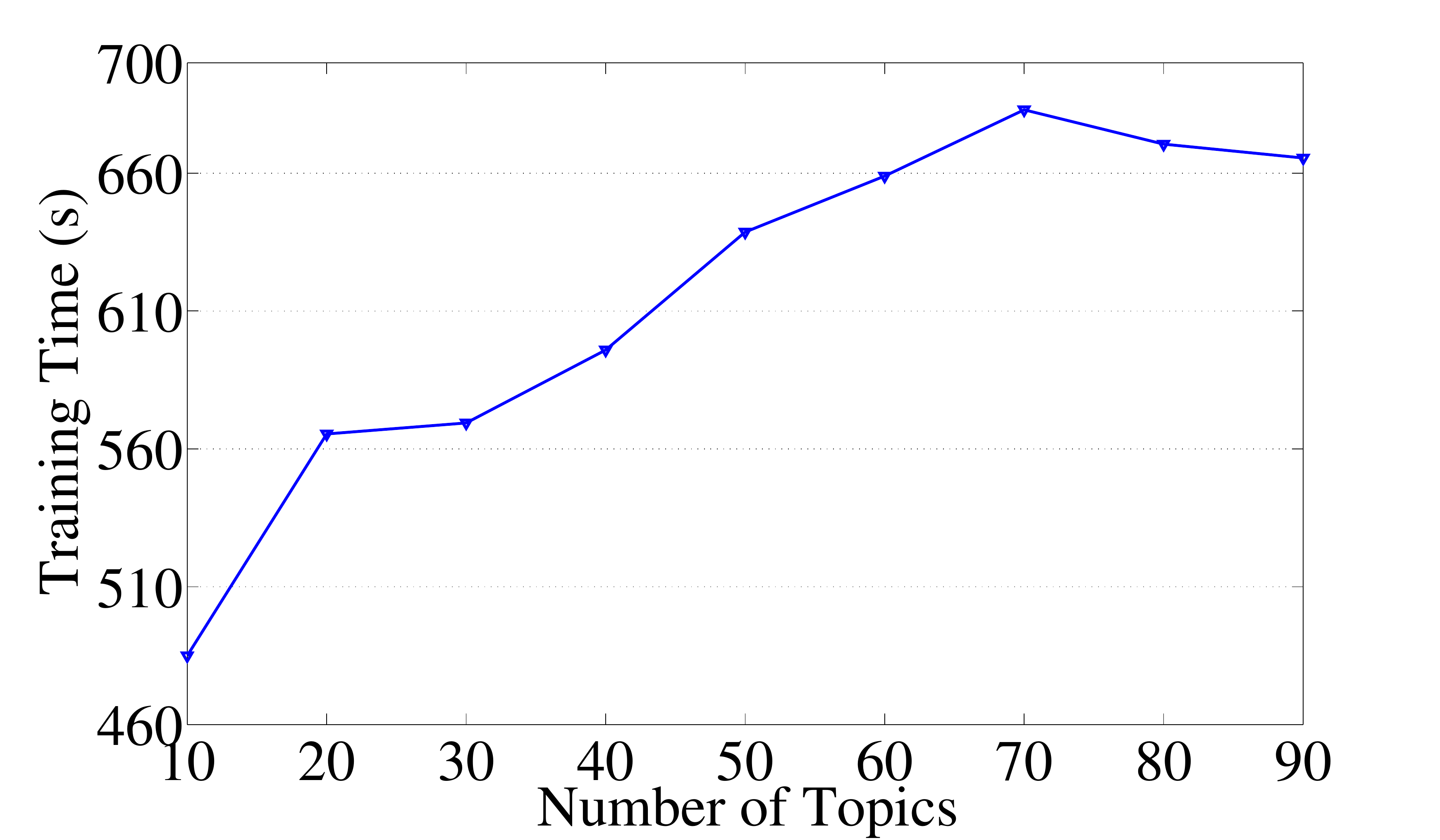}
\caption{Scalability vs. Number of Topics. As we increase the number of topics, the 
average training time also increases gradually.}
\label{fig:time-vs-topics}
\end{center}
\end{figure}

The number of topics is another important factor affecting the training time
(and hence the scalability) of the model. Figure~\ref{fig:time-vs-topics} shows
the average time for one iteration against different numbers of topics.  In this
experiment, we use $10\%$ data (over $40k$ documents) to train, and the time is
measured after model convergence. As the number of topics we want to model
increases, the training time for every iteration also increases, as
additional machines take up the additional load.

Ideally, these increases should be perfectly linear with the size of input
and/or number of topics. However, MapReduce framework involves machine cycle
scheduling, data load balance, and disk I/O operations between each
iteration. These factors highly rely on the underlying hardware and network.

The synchronization overhead of MapReduce is related to the number of keys
emitted by mappers and ability of the cluster to transmit and process these
data.  In Mr. LDA, every mapper emits $O(T_{d}K)$ messages, where $T_{d}$ is the
number of types in document $d$ and $K$ is the number of topics (in practice, it
could be less, as combiners can combine messages within mappers).  To
empirically validate this linear growth in both data size and the complexity of
the model, we ran Mr. LDA on the entire dataset with $100$ topics and $10$
topics. The intermediate data shuffled by the platform is approximately $10$
times larger -- $6.470$ GB ($466.60$ million records) for $100$ topics
vs. $646.79$ MB ($46.66$ million records) for $10$ topics. Combiners in both
cases help to reduce the intermediate data significantly. In the $100$ topics
scenario, combiners merge more than $16$ billion key value pairs at the mapper
side, whereas for $10$ topics case, they merge slightly more than $1.6$ billion
records.

To better test the scalability of our implementation, we further measure the
training time under different numbers of mappers.  Again, the training time is
measured over $20$ EM iterations and the number of topics is set to $10$. Total
number of input documents is $472525$.

\begin{figure}[htb]
\begin{center}
\includegraphics[width=0.9\linewidth]{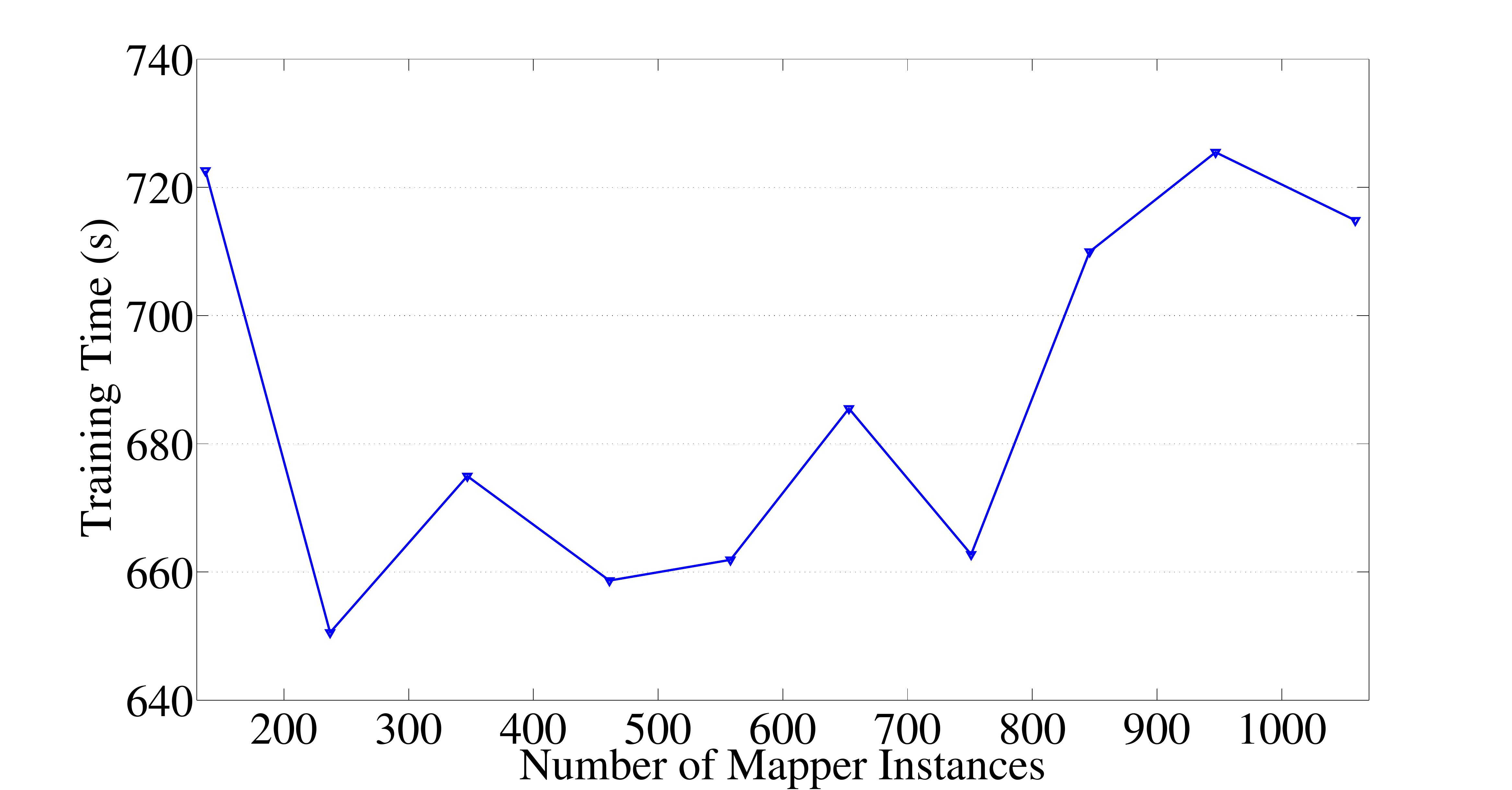}
\caption{Scalability vs. Number of Mapper Instances. The number of mapper
  instances represents a trade-off between number of processing units and
  network traffic due to intermediate data transfer. A larger number of mappers
  provides more computational resources, but also creates network
  congestion during system shuffling and sorting.}
\label{fig:training time scaling mapper}
\end{center}
\end{figure}

As illustrated in Figure~\ref{fig:training time scaling mapper}, we observe that
training time first decreases as we increase the number of mappers. Eventually,
for a fixed number of documents, adding additional mappers increases the
computation time.  This is because each mapper processes fewer documents but
still has fixed startup costs and because more mappers generate greater network
congestion (fewer opportunities to combine results). As with many MapReduce
algorithms, one must choose the correct amount of resources to solve a
problem.

\zkcomment{I indeed suggest remove this figure, as it might confuse the
  reviewer. Basically, we want to show that, number of mappers is greatly
  related to the size of input data, a larger number of mappers does not
  necessary ``scale'' the model, we should set it to a proper value. However, I
  think this diagram confuse the reviewers. shall we just show up to 800
  mappers, or just remove this figure out?}

While the number of reducers is important in general, the majority of the work
in Mr. LDA is done in the mappers. Therefore, the number of map tasks depends on
the size of the corpus; however, a single reducer, with a pass through the
vocabulary, is comparatively simple. As long as the number of reducer instances
is greater than the number of topics, reducers will not be a significant
bottleneck.

If we assume all the topics are used, which is generally true in LDA, we will
have subsets with approximately equal size. Hence, the entire workload will be
distributed somewhat evenly among all reducer instances. In this case, it is
unlikely that one reducer will delay the termination of the MapReduce step. This
is further assisted by the use of combiners, which can preemptively do some of
the reducers' work.

\subsection{Held-out Likelihood}
\label{sec:exp:likelihood}

Unlike the Gibbs sampling algorithms discussed in
Section~\ref{sec:scaling:gibbs}, which sacrifice the semantics of inference to
improve scalability, Mr. LDA's inference is identical to that conducted on a
single machine. Thus, there is no need to compare likelihood against
stand-alone implementations. However, It is useful to examine likelihood to
determine the number of iterations (and synchronizations) necessary for
inference.

\begin{figure}[htb]
\begin{center}
\includegraphics[width=0.9\linewidth]{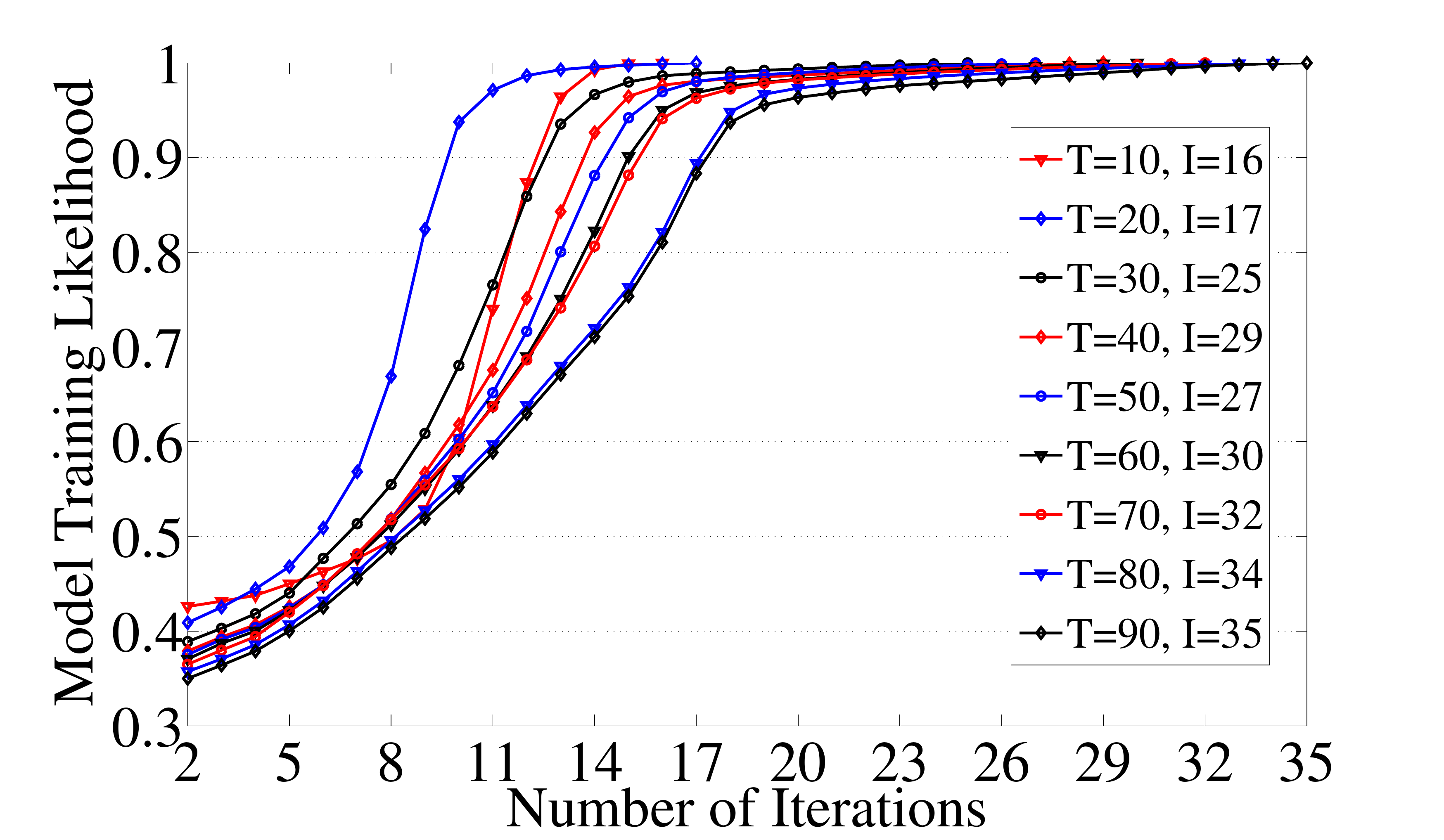}
\caption{Normalized Training Likelihood vs. Number of Topics. Here, the
  likelihood is scaled by the likelihood bound at convergence so that we can
  compare different number of topics.  The iteration at which the model
  converged is shown in the legend as $I$.  Generally, more topics require more
  iterations for the algorithm to converge, but the number of iterations is in
  the dozens rather than in the hundreds (as with Gibbs sampling).}
\label{fig:training-lhood}
\end{center}
\end{figure}

Figure~\ref{fig:training-lhood} shows the training likelihood against the number
of iterations.  To ease comparisons between different numbers of topics, the
likelihood has been divided by the final likelihood bound value.  The legend
shows the number of topics and number of iterations to converge. For example, if
we want to train $30$ topics on the training dataset, it would take $25$
iterations to converge. More complex models with more topics understandably
require more iterations to converge.  While this is common knowledge (and
independent of MapReduce), we stress this point because of its contrast with
Gibbs sampling, which typically takes hundreds of iterations or more to
converge~\cite{smola-10} with substantially more synchronizations required.
When the expensive computations can be easily parallelized, as is the case with
both Gibbs sampling and variational inference, one should try to minimize the
number of steps where an explicit synchronization must take place.

\hidetext{
Recall the results reported in Y!LDA
, in large cluster
environment, Gibbs sampling typically takes \textbf{hundreds} of iterations to
converge. While in our experiments, for a similar size of input documents,
variational approaches tend to take far less iterations, i.e. \textbf{dozens}
of. Even though the results are not directly comparable, it is a well
established statement that Gibbs sampling in general takes larger number of
iterations to converge. 

As we discussed in Section~\ref{} earlier, another inevitable drawback of Gibbs
sampling is that, in large scale settings, Gibbs sampling becomes
``constrained'' randomization (i.e., random seeds in pLDA~\cite{wang-09}, shared
state in Y!LDA~\cite{smola-10}). Recall the results reported for
Y!LDA~\cite[Figure 4]{smola-10}, with same settings, the likelihood for single
machine LDA is orders of magnitude higher than multi-machine LDA. Variational
inference, on the other hand, performs in an exactly same mechanism as in single
machine environment. It does not suffer from this problem as it does not rely on
state explore in MCMC process, and does not make any compromise or trade-offs in
likelihood.

\zkcomment{Jordan, would you please rephrase the above two paragraphs?
  \jbgcomment{shortened considerably, and pulled out salient points - I didn't
    repeat the constraints, as I think we already hit that.}}  }

\jbgcomment{It would be nice to have error bars on all of these graphs with
  random restarts and random scheduling (to vary both the initialization and
  vagaries of when it ran on the cluster, although the last one is only relevant
  for the timing information)}

\subsection{Informed Priors}
\label{sec:exp:prior}

\hidetext{
\begin{table*}
\caption{LIWC categories and their relevant words and word stems. Word stems are marked with asterisks.}
\center
\scriptsize{
\begin{tabular}{p{1.0cm} p{1.0cm} p{1.0cm} p{1.0cm} p{0.8cm} p{0.8cm} p{1.0cm} p{0.8cm} p{1.0cm} p{1.0cm} p{0.9cm} p{1.0cm}}
\hline
Affective Processes & Negative Emotions & Positive Emotions & Anxiety
& Anger & Sadness & Cognitive Processes & Insight & Causation & Discrepancy & Tentative & Certainty \\
\hline
abandon* & abandon* & accept & afraid & abuse* & abandon* & abandon* & accept & activat* & besides & allot & absolute\\
damn* & enrag* & freed* & alarm* & jealous* & ache* & conflict* & informs & affect & could & almost & absolutely \\
fume* & maddening & partie* & anguish* & abusi*	& aching & harness* & accepta* & affected & couldnt & alot & accura* \\
kindn* & snob*	& accepta* & anxi* & jerk & agoniz* & needed & inquir* & affecting & couldn't & ambigu* & all \\
privileg* & abuse* & freeing & apprehens* & aggravat* & agony & should've & accepted & affects & couldve & any & altogether \\
supporting & envie* & party* & asham* & jerked & alone & absolute & insight* & aggravat* & could've & anybod* & always \\
abuse* & madder & accepted & aversi* & aggress* & broke & confus* & accepting & allow* & desir* & anyhow & apparent \\
danger*	& sob & freely & avoid* & jerks & cried & hazie* & inspir* & attribut* & expect* & anyone* & assur* \\
fuming & abusi* & passion* & awkward* & kill* & cries & needing & accepts & based & hope & anything & certain* \\
\hline
\end{tabular}}
\label{tbl:liwc}
\end{table*}
}

In this set of experiment, we build the informed priors from
LIWC~\cite{pennebaker-99} dictionary as we discussed in
Section~\ref{sec:flexibility:informed prior}. Besides TREC dataset, we also used
the same informed prior on the BlogAuthorship corpus~\cite{koppel-06}, which
contains about 10 million blog posts from American users. In contrast to the
newswire-heavy TREC corpus, the BlogAuthorship corpus is more personal and
informal. Again, terms in fewer than $20$ documents are excluded, resulting
$53000$ types. Throughout the experiments, we set the number of topics to 100,
with $12$ guided by the informed prior.

The results are shown in Table~\ref{tbl:informed prior}. The prior acts as a
seed, causing words used in similar contexts to become part of the topic. This
is important for computational social scienctists who want to discover how an
abstract idea (represented by a set of words) is \emph{actually} expressed in a
corpus. For example, public news media (e.g., news articles like TREC) relates
positive emotions to entertainment, such as music, film and TV, whereas social
media (e.g., blog posts) relates it to religion. The \emph{Anxiety} topic in
news relates to middle east, but in blogs, it focuses on illness, e.g. bird flu. In
both corpora, \emph{Causation} was linked to science.

Using informed priors can discover radically different words. While LIWC is
designed for relatively formal writing, it can also discover Internet slang such
as ``lol'' (``laugh out loud'') in \emph{Affective Process} category. On the
other hand, some discovered topics do not have a clear relationship with the
initial LIWC categories, such as the abbreviations and acronyms in
\emph{Discrepancy} category.

\hidetext{}

\begin{table*}
\center
\scriptsize{
\begin{tabular}{p{0.5cm} | p{1.0cm} p{1.0cm} p{1.0cm} p{1.0cm} p{0.8cm} p{0.7cm} p{1.0cm} p{1.0cm} p{0.8cm} p{1.0cm} p{0.8cm} p{0.9cm}}
\hline
& Affective Processes & Negative Emotions & Positive Emotions & Anxiety & Anger & Sadness & Cognitive Processes & Insight & Causation & Discrepancy & Tentative & Certainty \\
\hline
\multirow{8}{*}{\begin{sideways}Output\end{sideways}\begin{sideways}from TREC\end{sideways}} & book & fire & film & al & polic & stock & coalit & un & technolog & pound & hotel & art \\
& life & hospit & music & arab & drug & cent & elect & bosnia & comput & share & travel & italian \\
& love & medic & play & israel & arrest & share & polit & serb & research & profit & fish & itali \\
& like & damag & entertain & palestinian & kill & index & conflict & bosnian & system & dividend & island & artist \\
& stori & patient & show & isra & prison & rose & anc & herzegovina & electron & group & wine & museum \\
& man & accid & tv & india & investig & close & think & croatian & scienc & uk & garden & paint \\
& write & death & calendar & peac & crime & fell & parliament & greek & test & pre & design & exhibit \\
& read & doctor & movie & islam & attack & profit & poland & yugoslavia & equip & trust & boat & opera \\
\hline
\multirow{8}{*}{\begin{sideways}Output\end{sideways} \begin{sideways}from Blog\end{sideways}} & easili & sorri & lord & bird & iraq & level & & god & system & sa & pretty & film \\
& dare & crappi & prayer & diseas & american & weight & & christian & http & ko & davida & actor \\
& truli & bullshit & pray & shi & countri & disord & & church & develop & ang & croydon & robert \\
& lol & goddamn & merci & infect & militari & moder & & jesus & program & pa & crossword & william \\
& needi & messi & etern & blood & nation & miseri & & christ & www & ako & chrono & truli \\
& jealousi & shitti & truli & snake & unit & lbs & & religion & web & en & jigsaw & director \\
& friendship & bitchi & humbl & anxieti & america & loneli & & faith & file & lang & 40th & charact \\
& betray & angri & god & creatur & force & pain & & cathol & servic & el & surrey & richard\\
\hline
\end{tabular}}
\caption{Twelve Topics Discovered from TREC (top) and BlogAuthorship (bottom) collection with LIWC-derived informed prior. The model associates TREC documents containing words like ``arab'', ``israel'', ``palestinian'' and ``peace'' with \emph{Anxiety}.  In the blog corpus, however, the model associates  words like ``iraq'',``america*'', ``militari'', ``unit'', and ``force'' with the \emph{Anger} category.}
\label{tbl:informed prior}
\end{table*}

\subsection{Polylingual LDA}
\label{sec:exp:poly}

As discussed in Section~\ref{sec:flexibility:polylingual}, Mr. LDA's modular
design allows us to consider models beyond vanilla LDA.  Using what we believe
is the first framework for variational inference for polylingual
LDA~\cite{mimno-09}, we fit $50$ topics to paired English and German Wikipedia
articles (approximately $500$k in each language). As before, we
ignore terms appearing in fewer than $20$ documents, resulting in $170$k English
word types and $210$k German word types.  While each pair of linked documents
shares a common subject (e.g. ``George Washington''), they are usually not
direct translations.  We let the program run for $33$ iterations with $100$
mappers and $50$ reducers; Table~\ref{tbl:exp:polylda} lists down some words
from a set of randomly chosen topics.

\zkcomment{the processing time was about $90$ minutes for first 4-5 iteration, then drop to 40-50 minutes for iteration 15+.}
\jbgcomment{Add in table and discussion for polylingual; don't discuss running time for this - the vocabulary is way too large
\zkcomment{yes, i agree. that relates to the tokenizer i used, it somehow did not fully stem the vocabularies, for example, you can see player, play and players appear together in one topic.}
\zkcomment{why not include running time? i put down a rough processing time for this. I think it is not that bad}}

\begin{table*}
\center
\scriptsize{
\begin{tabular}{c | p{0.9cm} p{1.1cm} p{1.1cm} p{1.1cm} p{0.6cm} p{1.0cm} p{1.0cm} p{1.1cm} p{1.0cm} p{0.6cm} p{0.9cm} p{1.0cm}}
\hline
\multirow{10}{*}{\begin{sideways}English\end{sideways}} & game & opera & greek & league & said & italian & soviet & french & japanese & album & york & professor\\
& games & musical & turkish & cup & family & church & political & france & japan & song & canada & berlin\\
& player & composer & region & club & could & pope & military & paris & australia & released & governor & lied\\
& players & orchestra & hugarian & played & childern & italy & union & russian & australian & songs & washington & germany\\
& released & piano & wine & football & death & catholic & russian & la & flag & single & president & von\\
& comics & works & hungary & games & father & bishop & power & le & zealand & hit & canadian & worked\\
& characters & symphony & greece & career & wrote & roman & israel & des & korea & top & john & studied\\
& character & instruments & turkey & game & mother & rome & empire & russia & kong & singer & served & published\\
& version & composers & ottoman & championship & never & st & republic & moscow & hong & love & house & received\\
& play & performed & romania & player & day & ii & country & du & korean & chart & county & member\\
%& video & instrument & romanian & match & wife & di & forces & louis & tokyo & albums & north & vienna\\
%& commic & dance & empire & win & died & saint & army & jean & sydney & singles & virginia & august\\
%& original & concert & bulgarian & final & left & king & communist & belgium & china & uk & senate & academy\\ 
%& manga & performance & bulgaria & teams & home & archbishop & led & belgian & red & records & carolina & 1933\\
%& ball & conductor & wines & scored & took & diocese & peace & les & arms & pop & congress & institute\\
\hline
\multirow{10}{*}{\begin{sideways}Germany\end{sideways}} & spiel & musik & ungarn & saison & frau & papst & regierung & paris & japan & album & new & berlin\\
& spieler & komponist & turkei & gewann & the & rom & republik & franzosischen & japanischen & the & staaten & universitat\\
& serie & oper & turkischen & spielte & familie & ii & sowjetunion & frankreich & australien & platz & usa & deutschen\\
& the & komponisten & griechenland & karriere & mutter & kirche & kam & la & japanische & song & vereinigten & professor\\
& erschien & werke & rumanien & fc & vater & di & krieg & franzosische & flagge & single & york & studierte\\
& gibt & orchester & ungarischen & spielen & leben & bishof & land & le & jap & lied & washington & leben\\
& commics & wiener & griechischen & wechselte & starb & italien & bevolkerung & franzosischer & australischen & titel & national & deutscher\\
& veroffentlic & komposition & istanbul & mannschaft & tod & italienischen & ende & russischen & neuseeland & erreichte & river & wien\\
& 2 & klavier & serbien & olympischen & kinder & konig & reich & moskau & tokio & erschien & county & arbeitete\\
& konnen & wien & osmanischen & platz & tochter & kloster & politischen & jean & sydney & a & gouverneur & erhielt\\
%& spiele & komponierte & jahrhundert & verein & kam & i & russland & pariser & japanischer & songs & john & august\\
%& dabei & kompositionen & bulgarien & league & sei & kaiser & staaten & pierre & china & erfolg & university & 1933\\
%& spielen & dirigent & budapest& 2008 & alter & maria & staat & et & wappen & you & amerikanischen & munchen\\
%& spiels & konservatorium & slowakei & kam & geboren & san & politische & les & australische & to & state & mitglied\\
%& ball & musiker & turkische & liga & wegen & erzbishof & israel & petersburg & japans & veroffentlicht & north & april\\
\hline
\end{tabular}}
\caption{Extracted Polylingual Topics from the Wikipedia Corpus. While topics
  are generally equivalent (e.g. on ``computer games'' or ``music''), some
  regional differences are expressed.  For example, the ``music'' topic in
  German has two words referring to ``Vienna'' (``wiener'' and ``wien''), while
  the corresponding concept in English does not appear until the
  $15^{\text{th}}$ position.}
\label{tbl:exp:polylda}
\end{table*}

\section{Conclusion and Future Work}
\label{sec:conc}

\jbgcomment{
Highlight new contributions:
\begin{enumerate}
	\item New distributed techniques for complete variational inference for LDA on MapReduce.  Unlike previous attempts this includes hyperparameter estimation (critical), multilingual support (nice), and informed priors (also nice)
	\item New variational inference for polylingual topic models, which had only been approached using Gibbs sampling
\end{enumerate}

Adding variational inference to the techniques that people can use for scaling up Bayesian models.  Important for both variety and for its versatility.
}

Understanding large text collections such as those generated via social media
requires algorithms that are unsupervised and scalable. In this paper, we
present Mr. LDA, which fulfils both of these requirements. Beyond text, LDA has
been successfully applied to other domains such as music~\cite{hu-09}, computer
vision~\cite{fergus-05}, biology~\cite{populationstructure}, and source
code~\cite{maskeri-08}. All of these domains struggle with the scale of data,
and Mr. LDA could help them better cope with large data.

Mr. LDA represents an alternative to the existing scalable mechanisms for
inference of topic models. Its design easily accomodates other extensions, as
we have demonstrated with the addition of informed priors and multilingual topic
modeling, and the ability of variational inference to support non-conjugate
distributions allows for the development of a broader class of models than could
be built with Gibbs samplers alone. Mr. LDA, however, would benefit from many of
the efficient, scalable datastructures that improved other scalable statistical
models~\cite{talbot-07}; incorporating these insights would further improve
performance and scalability.

While we focused on LDA, the approaches used here are applicable to many other
models. Variational inference is an attractive inference technique for the
MapReduce framework, as it allows the selection of a variational distribution
that breaks dependencies among variables to enforce consistency with the
computational constraints of MapReduce. Developing automatic ways to enforce
those computational constraints and then automatically derive
inference~\cite{winn-05} would allow for a greater variety of statistical models
to be learned efficiently in a parallel computing environment.

Variational inference is also attractive for its ability to handle online
updates. Mr. LDA could be extended to more efficiently handle online batches in
streaming inference~\cite{hoffman-10}, allowing for even larger document
collections to be quickly analyzed and understood.

\bibliographystyle{style/IEEEtran}
\bibliography{journal-abbrv,jbg}

\begin{thebibliography}{10}
\providecommand{\url}[1]{#1}
\csname url@samestyle\endcsname
\providecommand{\newblock}{\relax}
\providecommand{\bibinfo}[2]{#2}
\providecommand{\BIBentrySTDinterwordspacing}{\spaceskip=0pt\relax}
\providecommand{\BIBentryALTinterwordstretchfactor}{4}
\providecommand{\BIBentryALTinterwordspacing}{\spaceskip=\fontdimen2\font plus
\BIBentryALTinterwordstretchfactor\fontdimen3\font minus
  \fontdimen4\font\relax}
\providecommand{\BIBforeignlanguage}[2]{{%
\expandafter\ifx\csname l@#1\endcsname\relax
\typeout{** WARNING: IEEEtran.bst: No hyphenation pattern has been}%
\typeout{** loaded for the language `#1'. Using the pattern for}%
\typeout{** the default language instead.}%
\else
\language=\csname l@#1\endcsname
\fi
#2}}
\providecommand{\BIBdecl}{\relax}
\BIBdecl

\bibitem{blei-03}
D.~M. Blei, A.~Ng, and M.~Jordan, ``Latent {D}irichlet allocation,''
  \emph{JMLR}, vol.~3, pp. 993--1022, 2003.

\bibitem{fergus-05}
F.~Rob, L.~Fei-Fei, P.~Pietro, and Z.~Andrew, ``Learning object categories from
  {G}oogle's image search.'' in \emph{ICCV}, 2005.

\bibitem{wang-09b}
C.~Wang, D.~Blei, and L.~Fei-Fei, ``Simultaneous image classification and
  annotation,'' in \emph{CVPR}, 2009.

\bibitem{airoldi-08}
E.~M. Airoldi, D.~M. Blei, S.~E. Fienberg, and E.~P. Xing, ``Mixed membership
  stochastic blockmodels,'' \emph{JMLR}, vol.~9, pp. 1981--2014, 2008.

\bibitem{populationstructure}
D.~Falush, M.~Stephens, and J.~K. Pritchard, ``Inference of population
  structure using multilocus genotype data: linked loci and correlated allele
  frequencies.'' \emph{Genetics}, vol. 164, no.~4, pp. 1567--1587, 2003.

\bibitem{boyd-graber-09}
J.~Boyd-Graber and D.~M. Blei, ``Multilingual topic models for unaligned
  text,'' in \emph{UAI}, 2009.

\bibitem{griffiths-05}
T.~L. {Griffiths}, M.~{Steyvers}, D.~M. {Blei}, and J.~B. {Tenenbaum},
  ``Integrating topics and syntax,'' in \emph{NIPS}, 2005.

\bibitem{dean-04}
J.~Dean and S.~Ghemawat, ``{MapReduce}: Simplified data processing on large
  clusters,'' in \emph{OSDI}, San Francisco, California, 2004, pp. 137--150.

\bibitem{dyer-08}
C.~Dyer, A.~Cordova, A.~Mont, and J.~Lin, ``Fast, easy and cheap: Construction
  of statistical machine translation models with {MapReduce},'' in
  \emph{Workshop on SMT (ACL 2008)}, Columbus, Ohio, 2008.

\bibitem{brants-07}
T.~Brants, A.~C. Popat, P.~Xu, F.~J. Och, and J.~Dean, ``Large language models
  in machine translation,'' in \emph{EMNLP}, 2007.

\bibitem{cohen-09}
S.~B. Cohen and N.~A. Smith, ``Shared logistic normal distributions for soft
  parameter tying in unsupervised grammar induction,'' in \emph{NAACL}, 2009.

\bibitem{mimno-09}
D.~Mimno, H.~Wallach, J.~Naradowsky, D.~Smith, and A.~McCallum, ``Polylingual
  topic models,'' in \emph{EMNLP}, 2009, IR.

\bibitem{neal-93}
R.~M. Neal, ``Probabilistic inference using {M}arkov chain {M}onte {C}arlo
  methods,'' University of Toronto, Tech. Rep. CRG-TR-93-1, 1993.

\bibitem{robert-04}
C.~Robert and G.~Casella, \emph{Monte {C}arlo Statistical Methods}, ser.
  Springer Texts in Statistics.\hskip 1em plus 0.5em minus 0.4em\relax New
  York, NY: Springer-Verlag, 2004.

\bibitem{teh-06b}
Y.~W. Teh, ``A hierarchical {B}ayesian language model based on {P}itman-{Y}or
  processes,'' in \emph{ACL}, 2006.

\bibitem{griffiths-04}
T.~L. Griffiths and M.~Steyvers, ``Finding scientific topics,'' \emph{PNAS},
  vol. 101, no. Suppl 1, pp. 5228--5235, 2004.

\bibitem{finkel-07}
J.~R. Finkel, T.~Grenager, and C.~D. Manning, ``The infinite tree,'' in
  \emph{ACL}, 2007.

\bibitem{newman-08}
D.~Newman, A.~Asuncion, P.~Smyth, and M.~Welling, ``{Distributed Inference for
  Latent Dirichlet Allocation},'' in \emph{NIPS}, 2008.

\bibitem{yan-09}
F.~Yan, N.~Xu, and Y.~Qi, ``Parallel inference for latent dirichlet allocation
  on graphics processing units,'' in \emph{NIPS}, 2009, pp. 2134--2142.

\bibitem{wang-09}
Y.~Wang, H.~Bai, M.~Stanton, W.-Y. Chen, and E.~Y. Chang, ``{PLDA:} parallel
  latent {D}irichlet allocation for large-scale applications,'' in \emph{AAIM},
  2009.

\bibitem{smola-10}
A.~J. Smola and S.~Narayanamurthy, ``An architecture for parallel topic
  models,'' \emph{VLDB}, vol.~3, 2010.

\bibitem{jordan-99}
M.~I. Jordan, Z.~Ghahramani, T.~S. Jaakkola, and L.~K. Saul, ``An introduction
  to variational methods for graphical models,'' \emph{Machine Learning},
  vol.~37, no.~2, pp. 183--233, 1999.

\bibitem{wainwright-08}
M.~J. Wainwright and M.~I. Jordan, ``Graphical models, exponential families,
  and variational inference,'' \emph{Foundations and Trends in Machine
  Learning}, vol.~1, no. 1--2, pp. 1--305, 2008.

\bibitem{blei-05}
D.~M. Blei and M.~I. Jordan, ``Variational inference for {D}irichlet process
  mixtures,'' \emph{Journal of Bayesian Analysis}, vol.~1, no.~1, pp. 121--144,
  2005.

\bibitem{teh-06}
Y.~W. Teh, M.~I. Jordan, M.~J. Beal, and D.~M. Blei, ``Hierarchical {D}irichlet
  processes,'' \emph{JASA}, vol. 101, no. 476, pp. 1566--1581, 2006.

\bibitem{kurihara-07}
K.~Kurihara, M.~Welling, and N.~Vlassis, ``Accelerated variational {D}irichlet
  process mixtures,'' in \emph{NIPS}, Cambridge, MA, 2007.

\bibitem{wolfe-08}
J.~Wolfe, A.~Haghighi, and D.~Klein, ``Fully distributed {EM} for very large
  datasets,'' in \emph{ICML}, 2008, pp. 1184--1191.

\bibitem{nallapati-07}
R.~Nallapati, W.~Cohen, and J.~Lafferty, ``Parallelized variational {EM} for
  latent {D}irichlet allocation: An experimental evaluation of speed and
  scalability,'' in \emph{ICDMW}, 2007.

\bibitem{mahout}
A.~S. Foundation, I.~Drost, T.~Dunning, J.~Eastman, O.~Gospodnetic,
  G.~Ingersoll, J.~Mannix, S.~Owen, and K.~Wettin, ``Apache {Mahout},'' 2010,
  \url{http://mloss.org/software/view/144/}.

\bibitem{wallach-09b}
H.~Wallach, D.~Mimno, and A.~McCallum, ``Rethinking {LDA}: Why priors matter,''
  in \emph{NIPS}, 2009.

\bibitem{mallet}
A.~K. McCallum, ``Mallet: A machine learning for language toolkit,'' 2002,
  http://www.cs.umass.edu/~mccallum/mallet.

\bibitem{asuncion-09}
A.~Asuncion, P.~Smyth, and M.~Welling, ``Asynchronous distributed learning of
  topic models,'' in \emph{NIPS}, 2008.

\bibitem{white-10}
T.~White, \emph{Hadoop: The Definitive Guide (Second Edition)}, 2nd~ed.,
  M.~Loukides, Ed.\hskip 1em plus 0.5em minus 0.4em\relax O'Reilly, 2010.

\bibitem{lin-10}
J.~Lin and C.~Dyer, \emph{Data-Intensive Text Processing with MapReduce}, ser.
  Synthesis Lectures on Human Language Technologies.\hskip 1em plus 0.5em minus
  0.4em\relax Morgan {\&} Claypool Publishers, 2010.

\bibitem{lin-09}
C.~Lin and Y.~He, ``Joint sentiment/topic model for sentiment analysis,'' in
  \emph{CIKM}, 2009.

\bibitem{asuncion-09b}
A.~Asuncion, M.~Welling, P.~Smyth, and Y.~W. Teh, ``On smoothing and inference
  for topic models,'' in \emph{UAI}, 2009.

\bibitem{minka-00}
T.~P. Minka, ``Estimating a dirichlet distribution,'' Microsoft, Tech. Rep.,
  2000, http://research.microsoft.com/en-us/um/people/minka/papers/dirichlet/.

\bibitem{pennebaker-99}
J.~W. Pennebaker and M.~E. Francis, \emph{Linguistic Inquiry and Word Count},
  1st~ed.\hskip 1em plus 0.5em minus 0.4em\relax Lawrence Erlbaum, August 1999.

\bibitem{yao-09}
L.~Yao, D.~Mimno, and A.~McCallum, ``Efficient methods for topic model
  inference on streaming document collections,'' 2009.

\bibitem{blei-07b}
D.~M. Blei and J.~D. Mc{A}uliffe, ``Supervised topic models,'' in
  \emph{NIPS}.\hskip 1em plus 0.5em minus 0.4em\relax MIT Press, 2007.

\bibitem{boyd-graber-08}
J.~Boyd-Graber and D.~M. Blei, ``Syntactic topic models,'' in \emph{NIPS},
  2008.

\bibitem{trec-22}
NIST, ``Trec special database 22,'' 1994, http://www.nist.gov/srd/nistsd22.htm.

\bibitem{snowball}
M.~Porter and R.~Boulton, ``Snowball stemmer,'' 1970,
  http://snowball.tartarus.org/credits.php.

\bibitem{koppel-06}
M.~Koppel, J.~Schler, S.~Argamon, and J.~Pennebaker, ``Effects of age and
  gender on blogging,'' in \emph{In AAAI 2006 Symposium on Computational
  Approaches to Analysing Weblogs}, 2006.

\bibitem{hu-09}
D.~Hu and L.~K. Saul, ``A probabilistic model of unsupervised learning for
  musical-key profiles,'' in \emph{ISMIR}, 2009.

\bibitem{maskeri-08}
G.~Maskeri, S.~Sarkar, and K.~Heafield, ``Mining business topics in source code
  using latent dirichlet allocation,'' in \emph{ISEC}, 2008.

\bibitem{talbot-07}
D.~Talbot and M.~Osborne, ``Smoothed bloom filter language models: Tera-scale
  lms on the cheap,'' in \emph{ACL}, 2007, pp. 468--476.

\bibitem{winn-05}
J.~Winn and C.~M. Bishop, ``Variational message passing,'' \emph{JMLR}, vol.~6,
  pp. 661--694, 2005.

\bibitem{hoffman-10}
M.~Hoffman, D.~M. Blei, and F.~Bach, ``Online learning for latent dirichlet
  allocation,'' in \emph{NIPS}, 2010.

\end{thebibliography}

\end{document}